%% file: main.tex
\documentclass{article}

\usepackage{microtype}
\usepackage{graphicx}
\usepackage{subcaption}
\usepackage{booktabs} % for professional tables

\usepackage{hyperref}

\usepackage[accepted]{icml2026}

\usepackage{amsmath}
\usepackage{amssymb}
\usepackage{mathtools}
\usepackage{amsthm}

\usepackage[capitalize,noabbrev]{cleveref}

\theoremstyle{plain}
\newtheorem{theorem}{Theorem}[section]

\theoremstyle{definition}

\theoremstyle{remark}

\usepackage[textsize=tiny]{todonotes}

\input{header}

\newcommand\blfootnote[1]{%
  \begingroup
  \renewcommand\thefootnote{}\footnote{#1}%
  \addtocounter{footnote}{-1}%
  \endgroup
}

\icmltitlerunning{Retaining by Doing: The Role of On-Policy Data in Mitigating Forgetting}

\begin{document}

\twocolumn[
  \icmltitle{Retaining by Doing: The Role of On-Policy Data in Mitigating Forgetting}



  \icmlsetsymbol{equal}{*}

  \begin{icmlauthorlist}
    \icmlauthor{Howard Chen}{princeton}
    \icmlauthor{Noam Razin}{princeton}
    \icmlauthor{Karthik Narasimhan}{princeton}
    \icmlauthor{Danqi Chen}{princeton}
  \end{icmlauthorlist}

  \icmlaffiliation{princeton}{Princeton Language and Intelligence, Princeton University}

  \icmlcorrespondingauthor{Howard Chen}{howardchen@princeton.edu}
  \icmlcorrespondingauthor{Noam Razin}{noamrazin@princeton.edu}

  \icmlkeywords{Machine Learning, ICML}

  \vskip 0.3in
]



\printAffiliationsAndNotice{}  % no special notice (required even if empty)

\input{sections/abstract}
\input{sections/intro}

\input{sections/sft_vs_rl}
\input{sections/gaussian_sim_forget}

\input{sections/analysis}

\input{sections/related}
\input{sections/discussion}
\input{sections/ack}
\input{sections/impact}

\nocite{langley00}

\bibliography{ref}
\bibliographystyle{icml2026}

%%%%%%%%%%%%%%%%%%%%%%%%%%%%%%%%%%%%%%%%%%%%%%%%%%%%%%%%%%%%%%%%%%%%%%%%%%%%%%%
%%%%%%%%%%%%%%%%%%%%%%%%%%%%%%%%%%%%%%%%%%%%%%%%%%%%%%%%%%%%%%%%%%%%%%%%%%%%%%%
% APPENDIX
%%%%%%%%%%%%%%%%%%%%%%%%%%%%%%%%%%%%%%%%%%%%%%%%%%%%%%%%%%%%%%%%%%%%%%%%%%%%%%%
%%%%%%%%%%%%%%%%%%%%%%%%%%%%%%%%%%%%%%%%%%%%%%%%%%%%%%%%%%%%%%%%%%%%%%%%%%%%%%%
\newpage
\appendix
\onecolumn

\input{appendix}

%%%%%%%%%%%%%%%%%%%%%%%%%%%%%%%%%%%%%%%%%%%%%%%%%%%%%%%%%%%%%%%%%%%%%%%%%%%%%%%
%%%%%%%%%%%%%%%%%%%%%%%%%%%%%%%%%%%%%%%%%%%%%%%%%%%%%%%%%%%%%%%%%%%%%%%%%%%%%%%

\end{document}

%% file: header.tex
\usepackage{booktabs}
\usepackage{url}
\usepackage{times}
\usepackage{latexsym}
\usepackage{hyperref}
\usepackage{amsmath,amssymb}
\usepackage{graphicx}
\usepackage{tabularx}
\usepackage{multirow}
\usepackage{pifont}
\usepackage{soul}
\usepackage[toc,page]{appendix}
\usepackage{xcolor}
\usepackage{makecell}
\usepackage{xspace}
\usepackage{colortbl}
\usepackage{caption}
\usepackage{graphicx}
\usepackage{subcaption}
\usepackage{wrapfig}
\usepackage{enumitem}
\usepackage{cancel}
\usepackage{bm}

\usepackage{amsthm}

\newtheoremstyle{mystyle}
  {\topsep}   % Space above
  {\topsep}   % Space below
  {\itshape}  % Body font
  {}          % Indent amount
  {\bfseries} % Theorem head font (bold)
  {\textbf{.}}         % Punctuation after theorem head
  { }         % Space after theorem head
  {\thmname{#1}\thmnumber{ #2}\normalfont\thmnote{ (#3)}} % Theorem head spec

\theoremstyle{mystyle}
\usepackage{arydshln}

\usepackage{minibox}
\usepackage{tabularx,ragged2e}
\usepackage{tabularx} % in the preamble
\usepackage{algorithm}
\usepackage{bbm}
\usepackage{extarrows}

%% file: sections/abstract.tex
\begin{abstract}

Adapting language models (LMs) to new tasks via post-training carries the risk of degrading existing capabilities---a phenomenon classically known as \emph{catastrophic forgetting}.
%In this paper, we set out to identify guidelines to mitigate this phenomenon, by systematically comparing the forgetting patterns of two widely adopted post-training methods: supervised fine-tuning (SFT) and reinforcement learning (RL).
In this paper, toward identifying guidelines for mitigating this phenomenon, we systematically compare the forgetting patterns of two widely adopted post-training methods: supervised fine-tuning (SFT) and reinforcement learning (RL).
Our experiments reveal a consistent trend across LM families (Llama, Qwen) and tasks (instruction following, general knowledge, and arithmetic reasoning): RL leads to less forgetting than SFT while achieving comparable or higher target task performance.
% Supervised fine-tuning (SFT) and reinforcement learning (RL) are widely used during post-training for adapting language models (LMs) to new tasks.
% However, they may inadvertently erode existing capabilities---a phenomenon classically known as \emph{catastrophic forgetting}.
% Our work begins by systematically comparing SFT and RL across a wide range of tasks, including instruction following, general knowledge, and arithmetic reasoning, using different model scales and families.
% We reveal a consistent trend: RL leads to less forgetting than SFT while achieving comparable or higher target task performance.
To investigate the cause for this difference, we consider a simplified setting in which the LM is modeled as a mixture of two distributions, one corresponding to prior knowledge and the other to the target task.
%Perhaps counterintuitively
%\kn{why counterintuitive? maybe just remove the phrase in abstract since you need to explain?}
We identify that the \emph{mode-seeking} nature of RL, which stems from its use of \emph{on-policy} data, enables keeping prior knowledge intact when learning the target task.
% In contrast, the \emph{mode-covering} nature of SFT spreads more probability mass from the prior knowledge component.
We then verify this insight by demonstrating that the use on-policy data underlies the robustness of RL to forgetting in practical settings, as opposed to other algorithmic choices such as the KL regularization or advantage estimation.
Lastly, as a practical implication, our results highlight the potential of mitigating forgetting using \emph{approximately} on-policy data, which can be substantially more efficient to obtain than fully on-policy data.
\end{abstract}

%% file: sections/intro.tex
%\begin{flushright}
%\textit{Quote about action and memory.} --- by someone
%\end{flushright}

\section{Introduction}\label{sec:intro}

\input{figs/teaser}

Adapting language models (LMs) to new target tasks during post-training carries the risk of eroding previously acquired capabilities---a phenomenon known as \emph{catastrophic forgetting} \citep{mccloskey1989catastrophic,kirkpatrick2017overcoming}.
Such forgetting has been reported to occur when training LMs to follow instructions via supervised fine-tuning (SFT) \citep{luo2023empirical, shi2024continual, wu2024continual} or aligning them with human preferences via reinforcement learning (RL) \citep{bai2022training,ouyang2022training}.
\blfootnote{Code is available at: \url{https://github.com/princeton-pli/retaining-by-doing}}

%\danqi{Do you cite these two papers for RLHF, or you cite these two papers for showing there are forgetting patterns in RLHF? I remember Ouyang et al have some results on performance degradation and mix in pre-training data. Is there any other work showing forgetting in RL.}
%\danqi{Another question is that you are only considering RLVR, instead of RLHF, right? Make better it clearer from the begining.}
However, the understanding of how SFT and RL compare in terms of their susceptibility to forgetting remains limited.
%However, while practitioners have observed that RL can preserve capabilities better than SFT, the field lacks rigorous isolation of which algorithmic components drive this difference—with conflicting hypotheses attributing it to KL regularization, advantage estimation, or data distribution.
In this work, we systematically compare the forgetting patterns of SFT and RL in order to identify principled guidelines for mitigating forgetting in LM post-training.
We conduct a comprehensive study across instruction following, general knowledge, and arithmetic reasoning tasks, using Qwen 2.5 \citep{yang2024qwen2_5} and Llama 3 \citep{grattafiori2024llama3} models of up to 8B scale.
%\danqi{you really need to say qwen2.5 and llama3}
Our experiments reveal a consistent trend: SFT suffers from severe forgetting, whereas RL can achieve high target task performance without substantial forgetting (Figure~\ref{fig:gain_drop}). 
%\nr{Potentially provide concrete numbers here or maybe we can just refer to Figure~\ref{fig:gain_drop}, which is quite self-explanatory? Can help in emphasizing that this is a main contribution}\danqi{I agree. I still feel this is one of the major contributions.}

We then investigate the cause for the relative robustness of RL to forgetting.
At first glance, it may seem at odds with conventional wisdom.
Namely, minimizing the cross-entropy loss via SFT is equivalent to minimizing the \emph{forward KL} divergence with respect to the optimal policy, while maximizing the RL objective corresponds to minimizing the \emph{reverse KL} \citep{korbak2022distribution}.
Conventional wisdom presumes that the \emph{mode-seeking} nature of reverse KL enables faster learning of target distributions \citep{chan2022greedification,tajwar2024suboptimal} at the cost of losing coverage of old modes, while the \emph{mode-covering} forward KL should maintain probability mass across modes.
We reconcile this discrepancy by considering a simplified setting, where the target distribution is modeled as a mixture of two distributions: one representing the policy's prior knowledge and the other representing the target task.
We show that, if the initial policy is uni-modal (\emph{i.e.}, has a single mode), then SFT can in fact be more robust than RL to forgetting. However, if the initial policy is multi-modal (\emph{i.e.}, has multiple modes), which is arguably the case for practical LMs, then mode-seeking RL leads to less forgetting than mode-covering SFT; see Figure~\ref{fig:teaser} for an illustration.

The mode-seeking behavior of RL (\emph{i.e.}, its accordance with reverse KL minimization) stems from the usage of \emph{on-policy} data.
Through extensive ablations, we empirically verify that this property underlies the robustness of RL to forgetting, as opposed to other algorithmic choices such as the advantage estimation or the application of KL regularization.
Moreover, we explore what degree of on-policy data allows mitigating forgetting.
We find that for SFT, while generating data only from the initial policy is not enough, \emph{approximately on-policy} data generated at the start of each epoch can suffice for substantially reducing forgetting.
%\danqi{SFT or RL -- in general, it is not very clear that you are ablating this for SFT in this paragraph}
This suggests a practical guideline for LM post-training: leveraging on-policy data, potentially sampled asynchronously or at the start of each epoch for improved efficiency, can reduce unintended disruption of the model's existing capabilities.

%Overall, our work highlights that RL is more resilient to forgetting than SFT, offers intuition as to why the mode-seeking nature of RL can counterintuitively lead to less forgetting, and provides actionable insights for mitigating forgetting through the usage of (approximately) on-policy data.

To summarize, our main contributions are:
%\begin{itemize}[leftmargin=6mm, topsep=2pt, itemsep=7pt, parsep=0pt]
%    \item We demonstrate that RL is more robust to forgetting than SFT through extensive experiments on instruction following, general knowledge, and reasoning tasks, using LMs from different families and scales.
%    \item We provide intuition for why the mode-seeking nature of RL, which stems from its use of on-policy data, can counterintuitively lead to less forgetting than mode-covering SFT.
%    \item We corroborate this insight by demonstrating that the use of on-policy data underlies the robustness of RL to forgetting in practical settings, and highlight the potential of mitigating forgetting through approximately on-policy data, which can be substantially more efficient to obtain than fully on-policy data.
%\end{itemize}
\begin{itemize}[leftmargin=7mm, topsep=2pt, itemsep=7pt, parsep=0pt]
    \item We provide a systematic empirical comparison of forgetting between SFT and RL across tasks, model families, and scales, establishing that RL forgets significantly less than SFT.
    \item We identify on-policy data as the core factor behind RL's robustness to forgetting---ruling out alternative explanations such as KL regularization or advantage estimation---and provide intuition for why mode-seeking updates can counterintuitively preserve prior knowledge.
    \item We show that \textit{approximately} on-policy data, generated at each epoch instead of each step, can still benefit from substantial forgetting mitigation with a lower computational cost, offering practical guidance for efficient post-training. 
    %Despite the popularity of iterative training in practice, we are the first to establish its benefit in forgetting mitigation.
    %\item We show that \textit{approximately} on-policy data, generated at each epoch instead of each step, achieves substantial forgetting mitigation at lower computational cost; 
    %despite the popularity of iterative training in practice, we are the first to establish its benefit for mitigating forgetting as an alternative to full-fledged RL.
    %despite the popularity of iterative training in practice, to our knowledge, its benefit for mitigating forgetting has not been previously studied.

\end{itemize}

%% file: figs/teaser.tex
\begin{figure*}[h]
    \centering
    \includegraphics[width=0.8\textwidth]{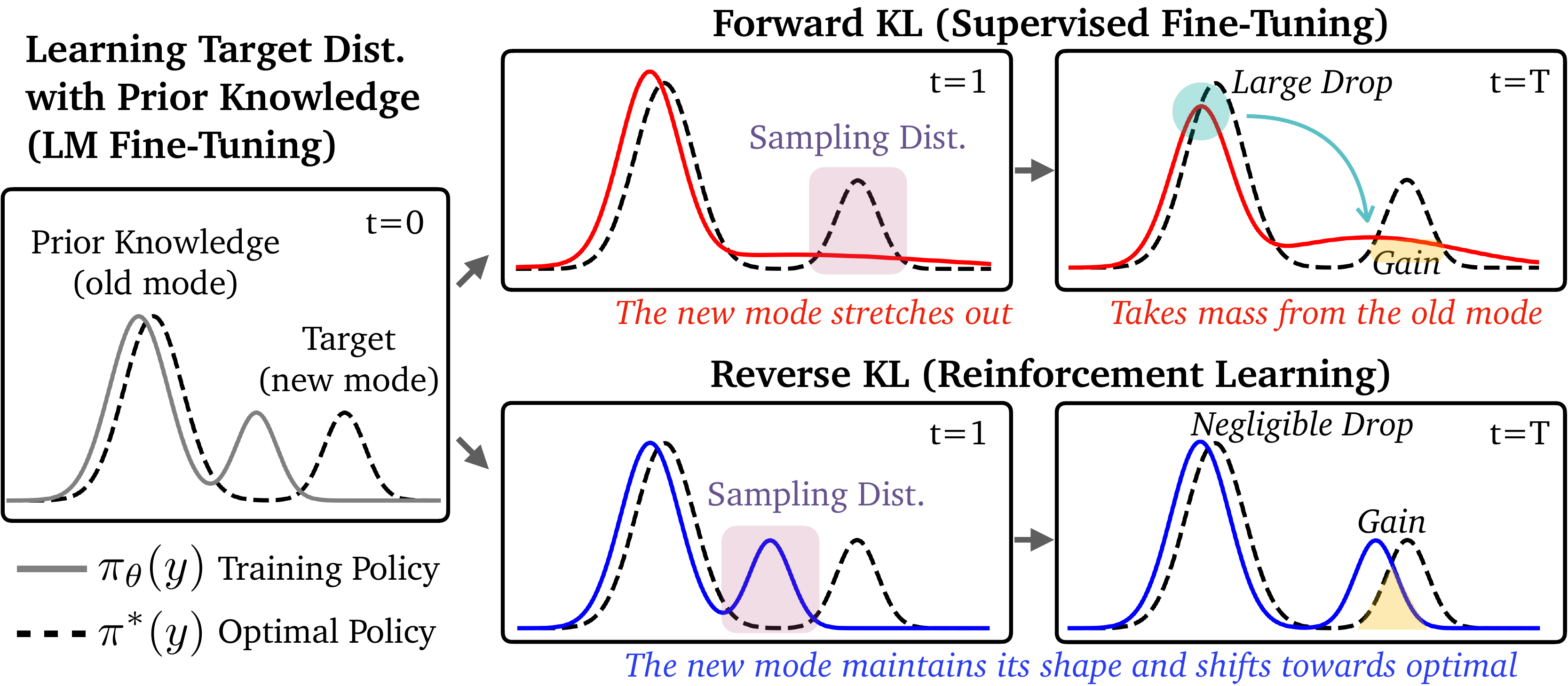}
    \caption{
        %\nr{In the blue text I believe it should be “The new mode maintains its shape..." (i.e., “its'' instead of “the''). Also, probably don't need to capitalize new mode and old mode in the red and blue text, for consistency.}
        Illustration of the forgetting dynamics for the forward KL objective, corresponding to SFT, and the reverse KL objective, corresponding to RL. 
        Left: we model LM post-training as a mixture of two modes. The “old'' mode represents prior knowledge and the “new'' mode represents a target task. Initially, the old mode of the training policy $\pi_\theta$ roughly matches the old mode of the optimal policy $\pi^*$, but its additional “new'' mode does not match the new target mode.
        The goal is for the training policy to match the optimal policy.
        Top right: minimizing forward KL first stretches the new mode of $\pi_\theta$ and then moves probability mass from the old mode to cover the target, leading to forgetting.
        Bottom right: in contrast, minimizing reverse KL maintains the shape of the old mode and covers the target distribution by shifting the new mode of $\pi_\theta$.
    }
    \label{fig:teaser}
\end{figure*}

%% file: sections/sft_vs_rl.tex
\section{Forgetting in LM Post-Training}\label{sec:sft_vs_rl}

We begin by introducing notation and the metrics used to measure forgetting.
Then, we empirically compare the forgetting patterns of supervised fine-tuning (SFT) and reinforcement learning (RL).
%We observe that SFT causes more forgetting than RL in the non-target domains. Comparing SFT variants, we note that SFT that learns from self-generated data forget less drastically compared to applying SFT on data generated by expert.

\subsection{Preliminaries}
\label{sec:prelim}
A language model (LM) is modeled by a policy $\pi_\theta(y \,|\, x)$, where the response $y$ is generated conditioned on the prompt $x$. For a target task $\mathcal{T}$, we denote the optimal policy by $\pi^*(\cdot \,|\, x)$. 
In SFT, the cross-entropy loss is minimized with respect to ground truth responses $y^*$ sampled from the optimal policy: 
$ \mathcal{L}_\mathrm{SFT}(\theta; x) := \sum\nolimits_{y} - \pi^*(y \,|\, x) \log \pi_\theta(y \,|\, x)$.
%\danqi{Can we say SFT is optimal policy?}
%\hc{SFT is not the optimal policy but it's learning from data sampled from the optimal policy.}
By contrast, in RL, the goal is to maximize the KL-regularized reward with respect to responses generated by the LM and a reward function $r(x, y) \in \{0, 1\}$\footnote{We use RL with verifiable reward (RLVR) throughout our experiments.}
%\danqi{Why do you assume reward is binary? Why not use a general form? Does it matter in terms of your experiments and analyses?},
\emph{i.e.}: $J_\mathrm{RL}(\theta; x) := \mathbb{E}_{y \sim \pi_\theta(\cdot \,|\, x)} [ r(x, y) ] - \beta \cdot \mathrm{KL}[\pi_\theta(\cdot \,|\, x) \, || \, \pi_{\theta_0}(\cdot \,|\, x) ]$, where $\beta > 0$ and $\pi_{\theta_0}$ is the initial policy.
%\danqi{Use log instead of ln.}
%\danqi{I think it is better to use two equation environments to denote LSFT and JRL}
%\hc{I feel like these objectives are pretty standard? Not sure if we need to use equation.}

% %A dataset for SFT training for task $\mathcal{T}$ is the collection of prompt-response pairs $D_\mathcal{T} = \{(x_i, y^*_i)\}$, and the dataset for RL is only the prompts $D_\mathcal{T} = \{x_i\}$.
% We denote an SFT dataset for a task $\mathcal{T}$, by $\mathcal{D}^{SFT}_{\mathcal{T}} = \{ (x_i, y^*_i)\}$ and the corresponding dataset  for RL, which consists of just the prompts, by $\mathcal{D}^{RL}_{\mathcal{T}} = \{ x_i \}$.

\input{figs/gain_drop}

\textbf{Forgetting and evaluation metrics.}
The initial policy $\pi_{\theta_0}$ is trained on a target task $\mathcal{T}$ for $T$ optimization steps, resulting in the trained policy $\pi_{\theta_T}$. This policy is evaluated using accuracy, which measures the fraction of correct outputs generated by $\pi_{\theta_T}$ for prompts associated with $\mathcal{T}$. 
We denote the accuracy of $\pi_{\theta_T}$ over $\mathcal{T}$ by $\mathcal{A}(\pi_{\theta_T}, \mathcal{T})$ and define the \textit{target task gain} as $\Delta_g := \mathcal{A}(\pi_{\theta_T}, \mathcal{T}) - \mathcal{A}(\pi_{\theta_0}, \mathcal{T})$.
%\danqi{you were using accuracy drop and accuracy gain in the introduction. just to be consistent, and i also there is no need to introduce notations in the intro.}
We quantify forgetting, based on on a collection of tasks $\{\mathcal{T}_j'\}_{j = 1}^M$, through the \textit{non-target tasks drop} $\smash{\Delta_d := \frac{1}{M} \sum\nolimits_{j = 1}^M \mathcal{A}(\pi_{\theta_0}, \mathcal{T}_j') - \mathcal{A}(\pi_{\theta_T}, \mathcal{T}_j')}$.
%\danqi{The reviewers will question whether the average makes sense. Why not just look at the drop over all the target tasks together?}
%\hc{it'd be really hard to present the results; I think this metric is not too out of the ordinary}
During post-training, the aim is to achieve high target task gain while minimizing as much as possible the non-target tasks drop.
For brevity, we will often refer to target task gain as \emph{gain} and to non-target tasks drop as \emph{drop}.

\subsection{Experimental Setup}\label{sec:sft_forgets_more_than_rl}

%\danqi{I think it is better to have a subsection called Experimental setup? so you can have paragraphs on 1) tasks 2) SFT/RL baselines; 3) model families. }
%\danqi{Maybe the title of this paragraph should be Target tasks.}
%\danqi{Results paragraph to Subsection: SFT forgets more than RL.}
%\hc{I feel it depends on if we want to further promote section 2.2, at the risk of emphasizing something that's not very surprising for the reader if they knew the other work or if we compare to them directly in the intro.}

\textbf{Target tasks and evaluation.}
We consider three tasks, covering different capabilities: IFEval \citep{zhou2023instruction, lambert2024tulu3} for instruction following, MMLU \citep{hendrycks2021measuring} for general knowledge, and Countdown \citep{tinyzero} for arithmetic reasoning.
These target task datasets are split into training set and evaluation set as described in Appendix~\ref{app:data}.
% We choose these dataset to cover a wide range of capabilities.
%\danqi{Countdown for reasoning is a bit overclaim. Let's downweigh it a bit. When people say reasoning, it is at least grade school tasks.}
%\danqi{Maybe you can add one sentence motivating why you have chosen these tasks.}
After training on one target task, we evaluate the model's performance on all the other tasks.
We additionally include the following non-target tasks: MATH \citep{hendrycks2021math}; two safety datasets, WildJailbreak \citep{jiang2024wildteaming} and WildGuardTest \citep{han2024wildguard}, since safety capabilities are often eroded through fine-tuning \citep{qi2024finetuning}, making them highly suitable for measuring forgetting.
%\danqi{Beter justify why you choose these datasets. Why do you want safety datasets here?}
In our RL experiments, correct generations are assigned a reward of $1$ and incorrect generations are assigned a reward of $0$.

\input{figs/small_lr_long}

\textbf{Models and baselines.}
We use instruct models from the Llama 3 \citep{grattafiori2024llama3} and Qwen 2.5 \citep{yang2024qwen2_5} families as the initial policies: Llama-3.2-1B-Instruct, Llama-3.1-8B-Instruct, Qwen-2.5-1.5B-Instruct, and Qwen-2.5-7B-Instruct. 
%\hc{cite}
%\danqi{No citations!}
We compare two SFT variants and one RL method: 
\textbf{1) SFT}, which uses responses generated by Llama-3.3-70B-Instruct as ground truth responses;
\textbf{2) Self-SFT}, which uses responses generated by the initial model (we keep only the correct responses based on the reward function \citep{zelikman2022star}); and \textbf{3) RL}---we use GRPO~\citep{shaoshuai2024deepseekmath}, a common algorithm for tasks with verifiable outputs.
For both SFT variants, the generated data was filtered using the reward function to include only examples with correct responses.
We use Self-SFT as a baseline to represent the typical first step in the post-training pipeline when human labels are absent \citep{dong2023raft, lambert2024tulu3}.
%\hc{justify Self-SFT and add citation}
%\danqi{any citations on self-SFT and expert-SFT?}
%\danqi{
%I think you should call expert SFT as SFT, and explain where (x, y) instruction comes from and put it as (1), and then you talk about
%(2) self-SFT, and cite relevant work to say why this is a meaningful/interesting baseline.
%(3) GRPO is too short, and you need to say why you chose GRPO.
%}
All models are trained for two epochs.
Additional implementation details are provided in Appendix~\ref{app:training_details}.

%\hc{mention policy gradient and GRPO}
%\hc{Make clear distinction between on-policy SFT vs off-policy SFT}

\subsection{Results: SFT Forgets More Than RL}\label{sec:sft_forgets_more_than_rl}

Figure~\ref{fig:gain_drop} compares the target task gain and non-target tasks drop of the SFT variants and RL.
We observe higher levels of forgetting in SFT compared to RL across datasets, model families, and sizes.
In particular, we find:
%For Self-SFT, achieving a similar target accuracy gain to GRPO induces a significantly larger drop on non-target tasks.
%While Expert-SFT can achieve a higher performance gain than GRPO on the instruction following task, it induces an even larger accuracy drop on non-target tasks relative to Self-SFT. 
%It is worth noting that, for SFT, a high learning rate is necessary to obtain high target performance, which often comes at the cost of severe forgetting. A smaller learning rate can avoid forgetting, yet fails to reach the same performance even when training for more epochs; see Figure~\ref{fig:small_lr_long}.
%Our experiments show a \textit{consistent tradeoff between target task performance and forgetting for SFT}.
%In contrast, GRPO improves target performance without a noticeable drop in non-target tasks.

\begin{itemize}[leftmargin=4mm]
  \item For Self-SFT, achieving a similar target accuracy gain to RL induces a significantly larger drop on non-target tasks.
  \item While SFT can achieve a higher performance gain than RL on the instruction following task, it induces an even larger drop on non-target tasks relative to Self-SFT.
  \item As shown in Figure~\ref{fig:small_lr_long}, a high learning rate is typically required to reach high target performance for SFT, often at the cost of severe forgetting; a smaller learning rate reduces forgetting but fails to reach the same target performance even with more epochs.
\end{itemize}
Overall, both SFT variants exhibit a consistent tradeoff between target performance and forgetting, whereas RL improves target performance without noticeable drops on non-target tasks.

%\danqi{
%Can you summarize the results in a few bullet points? It will make it easier to read?
%Also, why is expert SFT better than RL a lot on IFEval?
%When you describe the results here, refer to some particular tasks and numbers?
%I even didn't realize Figure 3 is part of the discussion. Use lists please!
%}
%\hc{Does referring to the numbers add much here? feels like the trend is clear without it?}
%\hc{Re: Expert-SFT better than RL - not very surprising since llama 70b's response is very well formatted and mimicing that style might be very helpful for the task. I don't think we need to explain this too much though.}

%% file: figs/gain_drop.tex
\begin{figure*}[!t]
    \centering
    \includegraphics[width=0.73\linewidth]{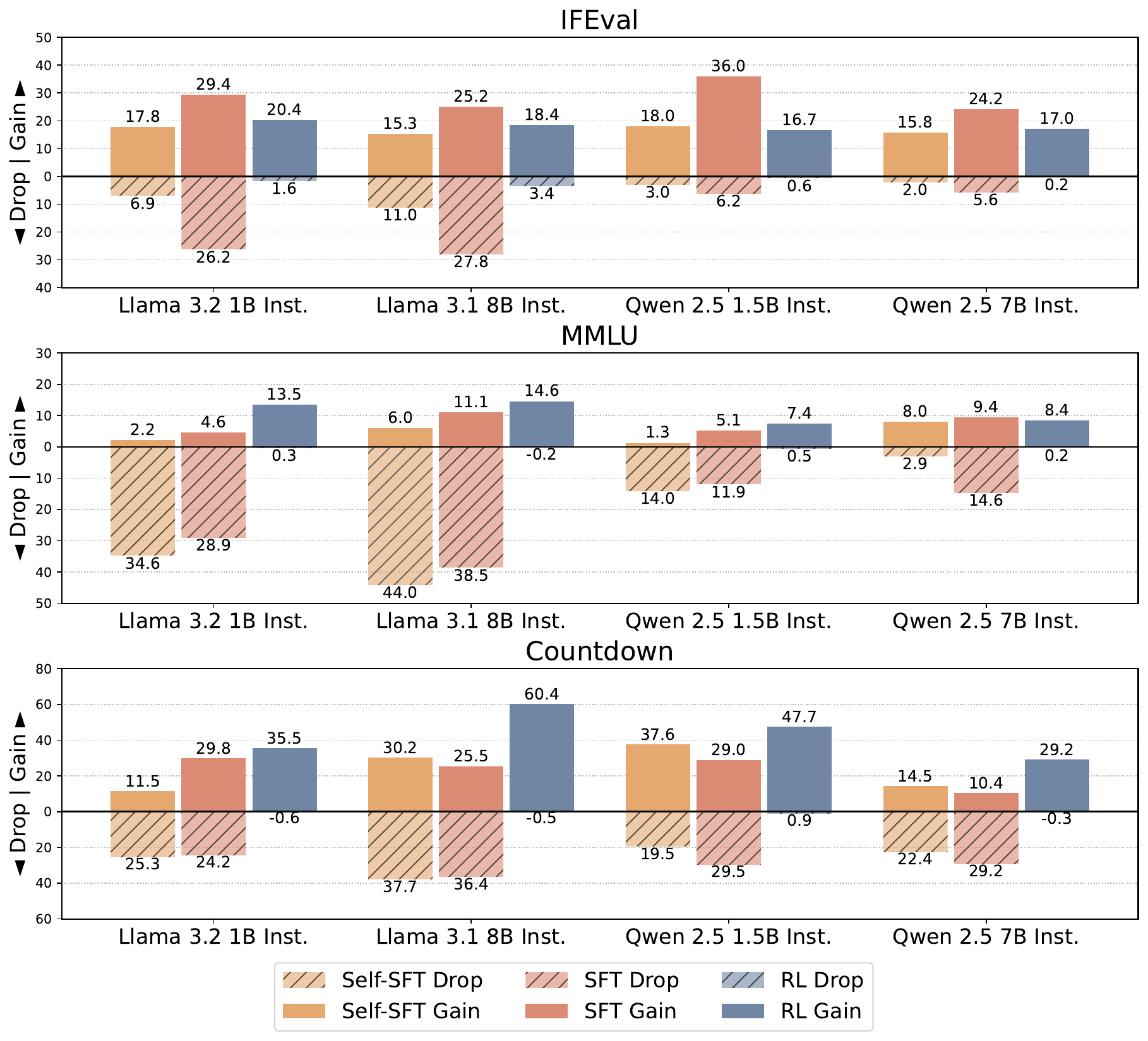}
    \caption{
    %\nr{Should add a sentence here summarizing the main takeaway from the plot. Can be in bold.}
      \textbf{SFT forgets more than RL across tasks and models.}
      We compare the \textbf{Gain} (solid bar) / \textbf{Drop} (shaded bar) across models and datasets for: \emph{(1)} Self-SFT, which uses data generated from the initial policy; \emph{(2)} SFT, which uses data generated by Llama-3.3-70B-Instruct; and \emph{(3)} RL (GRPO).
      Gain (higher better) represents the accuracy increase on the target task, while drop (lower better) represents the average accuracy decrease on non-target tasks. 
      %\danqi{you really really should label the models as llama3.1 llama 3.2 qwen2.5}
            %\danqi{Nit: instead of putting ``Gain / Drop'' as y label. I would put Gain with an uparrow to indicate the parts above the line, and Drop with a downarrow to indicate the parts below the line.}
            %\danqi{Add one sentence in the caption about the drop is measured as average of all non-target tasks, which equals to the other two tasks + two safety datasets?}
            %For presentation purposes, the drop shown in the plot 
      %\nr{There is an inconsistency between this plot, which shows drop in negative values (and so lower is actually not better) and the definition of drop, which is positive when you lose performance. Can explicitly say that for presentation purposes drop is presented as negative values, define drop differently, or perhaps add another y axis on the right with inverted ticks (positive on bottom and negative on top) labeled Drop and label the left axis only by Gain.}
    }
    \label{fig:gain_drop}
\end{figure*}

%% file: figs/small_lr_long.tex
\begin{figure*}[!t]
    \centering
    \includegraphics[width=0.95\linewidth]{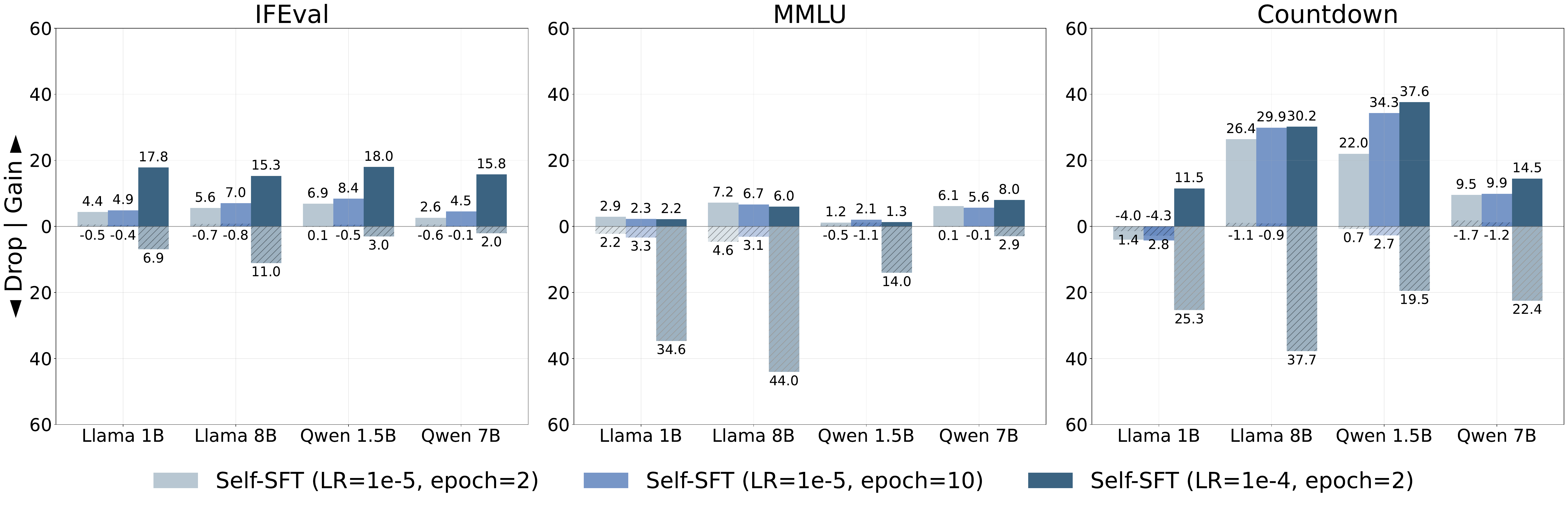}
    \caption{
      %\nr{Should add a sentence here summarizing the main takeaway from the plot. Can be in bold.}
      \textbf{SFT exhibits a tradeoff between target task performance and forgetting.}
      Comparison of Self-SFT runs with learning rates: $1e{-5}$ (small) and $1e{-4}$ (default), and training epochs ($2, 10$).
      %\nr{I believe this is actually Self-SFT here right? Worth mentioning in caption or legend that lr $1e-4$ is the default (i.e. same results as in Figure 2)}
      %\nr{There is a “ in the y-axis label that got in there by mistake.}
    }
    \label{fig:small_lr_long}
\end{figure*}

%% file: sections/gaussian_sim_forget.tex
\section{Understanding Forgetting Dynamics Through the Lens of KL}\label{sec:forgetting_dynamics}

%\danqi{is it better to say `KL Divergence' in the title?}

SFT and RL can be viewed as minimizing different directions of the KL divergence with respect to the optimal policy.
Specifically, as reviewed in \S\ref{sec:post_train_as_kl}, SFT corresponds to forward KL minimization while RL corresponds to reverse KL minimization. Intuitively, a \emph{mode-seeking} objective such as reverse KL should be more susceptible to forgetting: it moves probability mass quickly from one mode to another, whereas \emph{mode-covering} forward KL should better maintain probability mass on all modes. This intuition is invalidated in light of the evidence presented in \S\ref{sec:sft_forgets_more_than_rl}, showing that SFT causes more forgetting than RL. We address this discrepancy 
%\danqi{describe the differences between 3.2 and 3.3 briefly} 
through an empirical analysis of a simplified setting with univariate Gaussian distributions.
The analysis reveals that SFT can in fact lead to less forgetting than RL if the initial policy has a single mode (\S\ref{sec:gaussian_uni}).
However, in multi-modal scenarios that arguably mirror more closely LM fine-tuning, we show that the mode-seeking properties of RL result in higher degrees of robustness to forgetting (\S\ref{sec:sim_setup}).
% We use the analyses to establish the intuition that the mode-seeking objective can lead to less forgetting in the presence of multiple modes, analogous to fine-tuning on a pretrained LM.

\subsection{SFT and RL as KL Minimization}\label{sec:post_train_as_kl}

\textbf{SFT as forward KL minimization (mode-covering).} 
It is widely known that SFT is equivalent to minimizing the forward KL between the optimal and training policies since:
\begin{align*}
    \mathcal{L}_\mathrm{SFT}(\theta; x) &= \sum\nolimits_{y} - \pi^*(y \, | \, x) \log \pi_\theta(y \, | \, x) \\
    &= \mathrm{KL} \big[ \pi^*(\cdot \, | \, x) \, || \, \pi_\theta(\cdot \, | \, x) \big] + \mathcal{H}(\pi^* ( \cdot \, | \, x) )
    \text{\,,}
\end{align*}
where $\mathcal{H}(\pi^* ( \cdot \, | \, x) )$ is the entropy of $\pi^* (\cdot \, | \, x)$, which does not depend on $\pi_\theta$.
    
\textbf{RL as reverse KL minimization (mode-seeking).} 
The optimal policy for the KL-regularized RL objective (\S\ref{sec:prelim}) is given by $\pi^*(y \, | \, x) = \frac{1}{Z(x)} \pi_{\theta_0} (y \, | \, x)\exp ( r(x, y) / \beta )$ \citep{korbak2022distribution}, where $\pi_{\theta_0}$ is the initial policy, $Z(x) := \sum\nolimits_{y} \pi_{\theta_0} (y \, | \, x)\exp ( r(x, y) / \beta )$ is the partition function, and $\beta > 0$ is the KL regularization coefficient.
This implies that one can view the maximization of the RL objective as minimization of the reverse KL from $\pi^*$  since:
\begin{align*}
    &J_\mathrm{RL}(\theta; x) \\
    &= \mathbb{E}_{y \sim \pi_\theta(\cdot \, | \, x)} [ r(x, y) ] - \beta \, \mathrm{KL}[\pi_\theta(\cdot \, | \, x) \, || \, \pi_{\theta_0} (\cdot \, | \, x) ] \\
    &=  - \beta \cdot \mathrm{KL}[ \pi_\theta(\cdot \, | \, x) \, || \, \pi^*(\cdot \, | \, x)] +\beta \, \log Z(x)
    \text{\,,}
\end{align*}
where $\log Z(x)$ does not depend on $\pi_\theta$ (\emph{c.f.} \citet{korbak2022distribution,tajwar2024preference}).

%\subsection{Canonical mode-seeking vs. mode-covering forgetting dynamics.}
\subsection{Forward KL Forgets Less in a Uni-Modal Setting}\label{sec:gaussian_uni}

\input{figs/simulation_curves_no_prior}

%\nr{Preamble is not so clear right now. I recommend making it clear here in one way or another that in this section we: (1) show that forward KL can lead to less forgetting, matching intuition dicussed above, if we consider an initial policy with a uni-modal distribution; (2) we show that by modeling the policy as a univariate Gaussian distribution and target as a mixture of two univariate Gaussian distribution: an “old'' distribution corresponding to prior knowledge and a “new'' distribution corresponding to the target task; (3) in the next section we will see however that when going from uni-modal to a multi-modal initial policy, reverse KL counterintuitively result in less forgetting.}

%In this section, we show that forward KL (SFT) leads to less forgetting due to its mode-covering nature than the reverse KL (RL) counterpart when we consider a uni-modal training policy.
%We set the optimal policy to be a mixture of two univariate Gaussian distributions as an analogy to LM fine-tuning, where the ``old'' mode corresponds to the prior knowledge and the ``new'' mode corresponds to the target task.
%The result of this synthetic experiment matches the intuition that a mode-covering objective forgets less as discussed in the beginning of the section.
%In the next section, we will show that when expanding the uni-modal training policy to multi-modal, reverse KL counterintuitvely causes less forgetting.

In this section, we demonstrate that forward KL (SFT) leads to less forgetting than reverse KL (RL) under a uni-modal training policy. We model the optimal policy as a mixture of two univariate Gaussian distributions to mirror LM fine-tuning: an ``old'' mode that corresponds to prior knowledge and a ``new'' mode that represents the target task. As shown below, results in this setting align with the intuition stated at the beginning of the section, by which the mode-covering forward KL should forget less. However, in the next section we show that once the uni-modal training policy is expanded to a multi-modal one, reverse KL causes less forgetting. 

%The mode-seeking objective typically moves the probability mass away from the new mode. In contrast, the mode-covering objective stretches the probability mass to cover the new mode but does not move the mode as rapidly.
%o illustrate this, we simulate the learning dynamics between forward KL and reverse KL using uni-modal Gaussian as the training policy.

\textbf{Setup.} The optimal policy is modeled by an “old'' mode representing prior knowledge and a “new'' mode representing a target task:
%\nr{I'd replace all $x$ notation in the analysis with $y$ or some other letter to avoid confusion with prompts (AFAICT, the $x$ here are closer in meaning to responses $y$).}
%\nr{Also, I'd use $\theta^*$ instead of $w$ for consistency of notation.}
\begin{equation}
    \pi^*(y) = \alpha^* \cdot p_\mathrm{old}(y; \theta^*_\mathrm{old}) + (1-\alpha^*) \cdot p_\mathrm{new}(y; \theta^*_\mathrm{new})
    \text{\,,}
    \label{eq:bi_modal_optimal}
\end{equation}
where $\alpha^* \in (0, 1)$ and the distributions $p_\mathrm{old}$ and $p_\mathrm{new}$ are univariate Gaussians with means and standard deviations given by $\theta^*_\mathrm{old} = ( \mu^*_\mathrm{old}, \sigma^*_\mathrm{old} )$ and $\theta^*_\mathrm{new} = ( \mu^*_\mathrm{new}, \sigma^*_\mathrm{new} )$, respectively.
%\nr{Need to define here that $p_{old}$ and $p_{new}$ are univariate Gaussians, currently it pops out of nowhere, and I'd use $m$ or $\mu$ for mean and $s$ or $\sigma$ for standard deviation (using $s$ for mean and $m$ for standard deviation is unnecessarily confusing).}
In this section, the training policy $\pi_\theta$ is modeled as a univariate Gaussian with trainable mean $\mu$ and standard deviation $\sigma$, \emph{i.e.}, $\theta = ( \mu, \sigma )$.
%We define the gain and drop accuracy as the difference in the \textbf{overlapping area}\footnote{The area of overlap can be expressed as the Total Variation metric. We provide full details in Appendix~\ref{app:tv}} between the training policy and the optimal policy. 
We define the target task gain and non-target tasks drop as the change in \emph{overlap area}\footnote{
The overlap area can be formulated via the total variation distance; see Appendix~\ref{app:tv}.
} between the training policy and the modes of the optimal policy.
Concretely, the overlap area for the old and new modes is defined as:
\begin{align}
S_\mathrm{old}(\theta) &:= \frac{\int_{-\infty}^\infty \min \left \{ \alpha^* p_\mathrm{old}(y), \pi_\theta(y) \right \} dy}{ \alpha^* } \notag \\ 
S_\mathrm{new}(\theta) &:= \frac{\int_{-\infty}^\infty \min \left \{ (1 - \alpha^*) p_\mathrm{new}(y), \pi_\theta(y) \right \}  dy}{1 - \alpha^*} \text{.}
\label{eq:uni_overlap_def}
\end{align}
Notice that $S_{\mathrm{old}} (\theta), S_{\mathrm{new}} (\theta) \in [0, 1]$.
%\nr{Isn't the denominator in $S_{\mathrm{old}}$ just $\lambda$? I replaced it, if I missed anything LMK}
The target task gain at training step $T$ is accordingly defined by $\Delta_g := S_\mathrm{new}(\theta_T) - S_\mathrm{new}(\theta_0)$ and the non-target tasks drop is $\Delta_d  := S_\mathrm{old}(\theta_0) - S_\mathrm{old}(\theta_T)$.
We initialize the training policy $\pi_\theta$ such that it covers the mode of $\pi^*$ corresponding to $p_{\mathrm{old}}$, and compare minimizing the forward and reverse KL objectives (defined in \S\ref{sec:post_train_as_kl}) with respect to $p_\mathrm{new}$ in terms of their gain-drop tradeoff. The parameters in $\theta$ are updated through sample-based gradients, where for forward KL data is sampled from $p_\mathrm{new}$ and for reverse KL it is sampled from $\pi_\theta$.
%\hc{Should we provide exact the objectives here?}
%\nr{Since the objectives are defined in the subsection above, not sure it is necessary. Can refer to them in parenthesis maybe?}
See Appendix~\ref{app:simu_details} for additional implementation details.

\textbf{Results.} Figure~\ref{fig:simulation_curves_no_prior} shows the forgetting patterns of forward and reverse KL. To reach a target task gain of $0.9$, forward KL results in a non-target tasks drop of $0.64$ while reverse KL leads to a larger drop of $0.7$. 
This matches common intuition: the mode-covering forward KL stretches the probability mass to cover the new mode while retaining more mass on the old mode compared to the mode-seeking reverse KL.
That is, in this setting, forward KL causes less forgetting than reverse~KL.

% The result matches common intuition: the mode-covering forward KL objective stretches the probability mass to cover the new mode but does not move away quickly, while the mode-seeking reverse KL moves the entire mass away, causing more forgetting.
%\nr{I'd add a sentence with the takeaway from this here.}
%\hc{cite Deep-RL forgetting papers}

\subsection{Reverse KL Forgets Less in a Multi-modal Setting}\label{sec:sim_setup}

\input{figs/simulation_curves}

%\hc{demote the theory/proof and clarify the examples better}

%\paragraph{Mirroring LM fine-tuning using a steady prior mode.}
%\nr{Need to polish this paragraph. I'd first say that in Section~\ref{sec:gaussian_uni} we saw that when the initial policy is uni-modal, the mode-covering properties of forward KL, which corresponds to SFT, can lead to less forgetting than reverse KL, which corresponds to RL, in accordance with intuition. However, the experiments of Section~\ref{sec:sft_vs_rl} demonstrated that in practical LM fine-tuning settings RL is more resilient to forgetting. Then, say that in this section we resolve this discrepancy by showing that if one considers an initial policy with multiple modes, which arguably matches practical settings more closely, mode-seeking reverse KL leads to less forgetting.}

We showed in \S\ref{sec:gaussian_uni} that the mode-covering properties of forward KL (SFT) lead to less forgetting than reverse KL (RL) when the initial training policy is uni-modal.
This stands in contrast to the experiments of \S\ref{sec:sft_forgets_more_than_rl}, which show that in practical LM post-training settings, RL is more resilient to forgetting.
In this section, we reconcile this discrepancy by showing that when we allow the initial training policy to have multiple modes, arguably a closer match to practice, the mode-seeking reverse KL results in less forgetting.

%We construct a setting where the training policy is modeled by two modes: a slow-moving left mode to mimic the probability mass of the pretrained knowledge, and a fast-moving right mode to represent the probability mass that can be allocated to the new task. 
%This setup mirrors the LM fine-tuning process more closely than the uni-modal setting, and we show that in the presence of the slow-moving left mode, the mode-seeking reverse KL leads to less forgetting.

\input{figs/kl_ablations}

%Inspired by \citep{kotha2024understanding}, we construct a mixture of two distributions: one distribution representing the existing capabilities (denoted ``old''), and the other mode represents the distribution of the new capability to acquire (denoted ``new''). 
\textbf{Setup.}
We consider the setup of \S\ref{sec:gaussian_uni}, where the optimal policy is modeled as a mixture of two Gaussian distributions (Equation~\ref{eq:bi_modal_optimal}).
Instead of modeling the training policy as a uni-modal Gaussian, we now model it as a bi-modal distribution:
%\nr{Worth being consistent with notation. I'd use $\pi_\theta$ and define $\theta = (\alpha, \theta_{\mathrm{old}}, \theta_{\mathrm{new}})$}
\begin{equation}
    \pi_\theta(y) = \alpha \cdot q_\mathrm{old}(y; \theta_\mathrm{old}) + (1-\alpha) \cdot q_\mathrm{new}(y; \theta_\mathrm{new})
    \text{\,,}
    \label{eq:bi_modal_learning}
\end{equation}
where $\theta = (\alpha, \theta_\mathrm{old}, \theta_\mathrm{new})$ is the trainable parameters, with $\alpha \in [0, 1]$ being a mixture weighting, $\theta_\mathrm{old} = ( \mu_\mathrm{old}, \sigma_\mathrm{old} )$ defining the mean and standard deviation of a univariate Gaussian $q_\mathrm{old}$, and $\theta_\mathrm{new} = (\mu_\mathrm{new}, \sigma_\mathrm{new})$ similarly defining a univariate Gaussian $q_\mathrm{new}$.
We initialize the training policy $\pi_\theta$ such that $q_{\mathrm{old}}$ roughly covers the mode of $\pi^*$ corresponding to $p_{\mathrm{old}}$ and, as in \S\ref{sec:gaussian_uni}, compare the gain-drop tradeoffs exhibited by forward and reverse KL minimization with respect to $p_{\mathrm{new}}$.
See Appendix~\ref{app:simu_details} for additional implementation details.

\textbf{Results.}

Figure~\ref{fig:simulation_curves} shows that
%, with a high learning rate, 
achieving a target task gain of $0.9$ with forward KL causes severe forgetting---the area overlap with $p_\mathrm{old}$ drops by $0.12$.
%Reducing the learning rate can mitigate forgetting of the old mode, but leads to failure in learning the target $p_\mathrm{new}$. 
By contrast, reverse KL shifts $q_\mathrm{new}$ toward $p_\mathrm{new}$ while largely keeping the old mode intact.
This simulation demonstrates that, for bi-modal policies, reverse KL can match a new target mode without redistributing probability mass from a mode that represents prior knowledge.

%% file: figs/simulation_curves_no_prior.tex
\begin{figure}[h]
    \centering    \includegraphics[width=1\columnwidth]{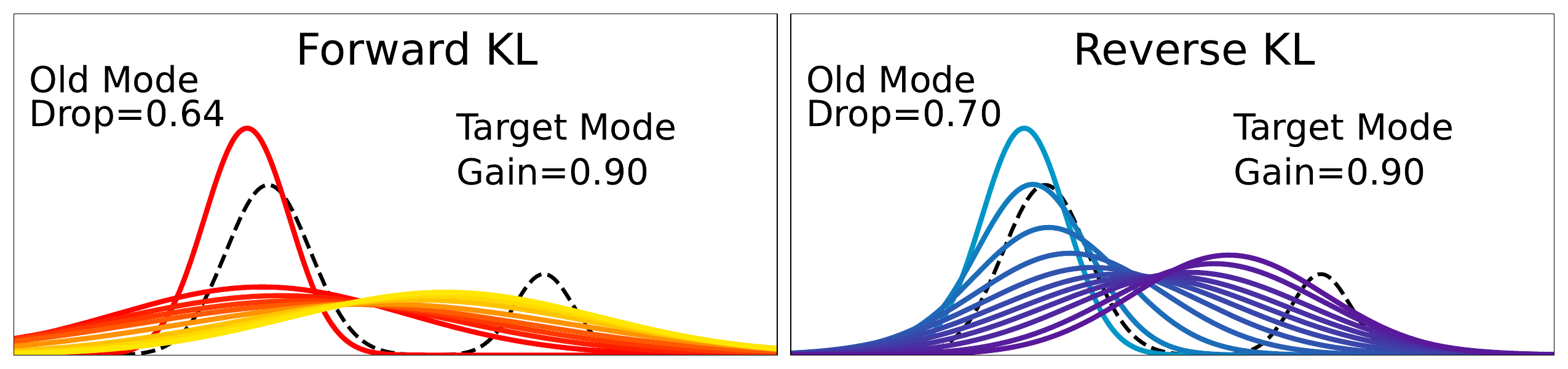}
    \caption{
        %\nr{1. I'd add a sentence here with the main takeaway from the plot, or rephrase the first sentence to also include the takeaway (i.e., something along the line that forward KL can lead to less forgetting when the initial policy is uni-modal)}
        %\nr{2. I'd add write in the titles Forward KL (SFT) and Reverse KL (RL), similarly to Figure~1. Also, it would help cosmetically to increase the margin between the title and top border of the figure.}
        \textbf{Forward KL (SFT) with uni-modal training policy forgets less than reverse KL (RL).}
        Learning and forgetting dynamics of forward KL (left) and reverse KL (right). 
        Dashed lines represent the modes of the optimal policy: $p_\text{old}$ (left) and $p_\text{new}$ (right).
        For forward KL, the data is sampled from the target mode $p_\text{new}$; the curve goes from red to yellow as training progresses. For reverse KL, the data is sampled from $\pi_\theta$; the curve goes from blue to purple.
        Forgetting corresponds to the decrease of overlap on the left mode and learning a new target task corresponds to the increase in overlap on the right mode.
    }
    \label{fig:simulation_curves_no_prior}
\end{figure}

%% file: figs/simulation_curves.tex
\begin{figure}[h]
    \centering
    \includegraphics[width=1\columnwidth]{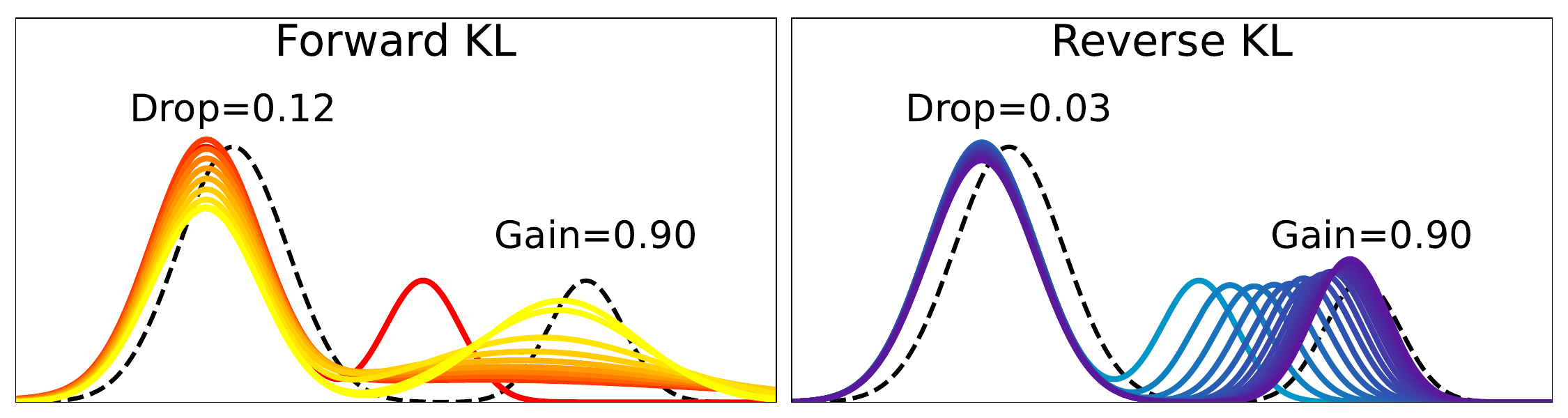}
    \caption{
        %\nr{I'd add a sentence here with the main takeaway from the plot, or rephrase the first sentence to also include the takeaway, and write in the titles Forward KL (SFT) and Reverse KL (RL), similarly to Figure~1}
        \textbf{Reverse KL (RL) with multi-modal training policy forgets less than forward KL (SFT).}
        Learning and forgetting patterns of forward KL (left)
        %with different high ($0.15$) and low ($0.01$) learning rates (left and middle) 
        and reverse KL (right).
        Dashed lines represent the modes of the optimal policy: $p_\text{old}$ (left) and $p_\text{new}$ (right). 
        For forward KL, the data is sampled from the target mode $p_\text{new}$; the curve goes from red to yellow as training progresses. For reverse KL, the data is sampled from $\pi_\theta$; the curve goes from blue to purple.
        Forgetting corresponds to the decrease of overlap on the left mode and learning a new task is the increase in overlap on the right mode.
    }
    \label{fig:simulation_curves}
\end{figure}

%% file: figs/kl_ablations.tex
\begin{figure*}[h]
    \centering
    \includegraphics[width=0.95\textwidth]{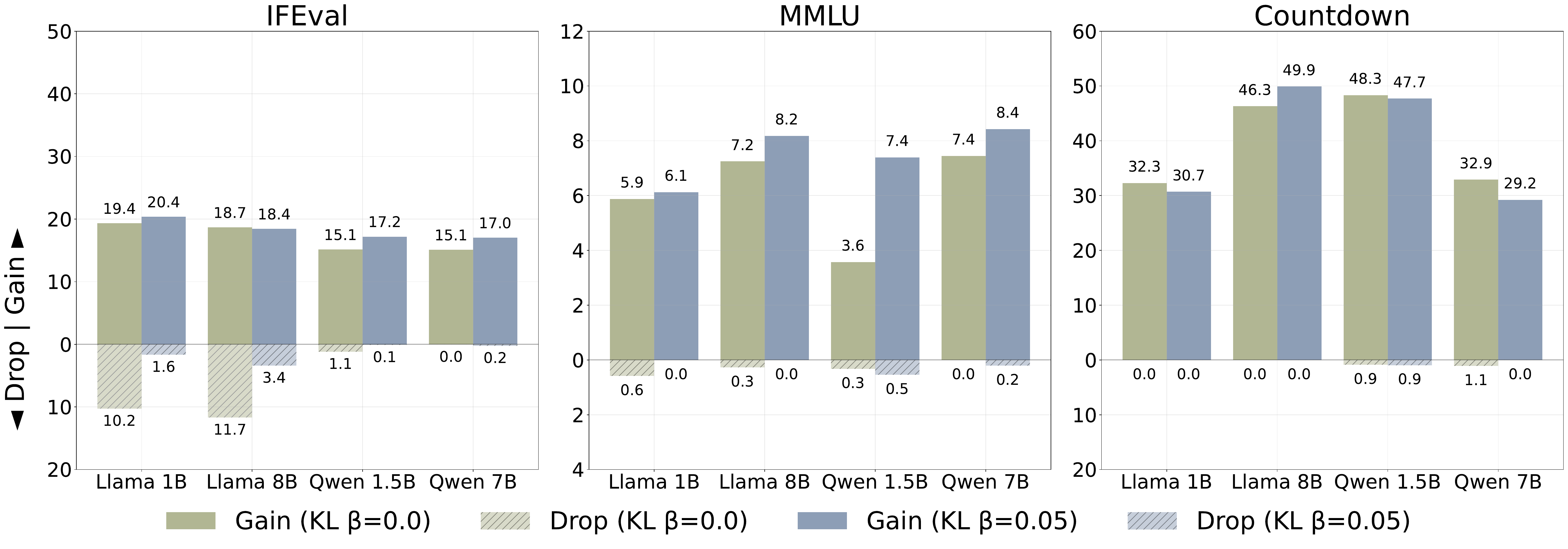}
    \caption{
        %\nr{I'd add a sentence here with the main takeaway from the plot, or rephrase the first sentence to also include the takeaway}
         %\nr{\color{red}{y-axis labeling is inconsistent with Figure~\ref{fig:gain_drop}}}
        \textbf{KL regularization is not a major contributor to RL's lesser degree of forgetting.}
        Comparison of GRPO with KL regularization ($\beta=0.05$) and without it ($\beta = 0.0$).
        Except for the Llama model family and IFEval target task, the non-regularized GRPO achieves a similar gain-drop tradeoff as regularized GRPO.
    }
    \label{fig:kl_ablations}
\end{figure*}

%% file: sections/analysis.tex
\section{Learning from On-Policy Data Mitigates Forgetting}\label{sec:analysis}

%\hc{
%Organize better so that it clearly makes the points:
%\begin{itemize}
%    \item On-policy data is responsible for mitigating forgetting (as opposed to advantage estimate or regularization). Can do this perhaps by first giving the REINFORCE and KL regularization experiments.
%    \item Then, say that on-policy data can be expensive and that this motivates exploring approximate on-policy data and give the relevant experiments
%\end{itemize}
%}

%\nr{
%I believe the positioning of this section should be refactored. It should contain two subsections.

%The first conveys that on-policy data can mitigate forgetting, and that also it can be approximate. We should emphasize the contrast from the results of \ref{fig:gain_drop}, where Self-SFT suffers from forgetting. So the results overall show that sampling once from the initial policy is not enough, but sampling each epoch can be enough. This then brings about a practical suggestions regarding mitigating forgetting more efficiently than completely on-policy updates.

%Second subsection shows that KL regularization is not responsible for the robustness of RL to forgetting, and hopefully shows some evidence against the main claim of \cite{shenfeld2025rlsrazor} (perhaps results of Figure~\ref{fig:gain_drop} will also provide some counter-evidence).
%}
%\nr{Candidate subsection titles: - Approximately on-policy data can suffice for mitigating forgetting OR On-Policy Data Mitigates Forgetting and Can be Approximate (not polished enough, but this is the rough idea) - KL Regularization Does Not Explain Robustness to Forgetting}

The experiments of \S\ref{sec:sft_vs_rl} demonstrated that RL causes less forgetting than SFT.
By considering a simplified setting in \S\ref{sec:forgetting_dynamics}, we identified that the mode-seeking behavior of RL, which stems from its usage of on-policy data, may underlie its robustness to forgetting.
We now verify this prospect by demonstrating in practical settings that the robustness of RL to forgetting indeed arises from its use of on-policy data, as opposed to other algorithmic choices such as KL regularization or an advantage estimator (\S\ref{sec:on_policy_explains}).
We then explore the following natural question: \emph{what degree of on-policy data allows mitigating forgetting?}
As evident from the results of Self-SFT in Figure~\ref{sec:sft_vs_rl}, generating data only from the initial policy is not enough.
However, we show that SFT with approximately on-policy data, generated at every epoch or with on-policy traces produced by RL, can suffice for substantially reducing forgetting (\S\ref{sec:approx_on_policy}).
This highlights the potential of mitigating forgetting through approximately on-policy data, which can be substantially more efficient to obtain than fully on-policy data.
% \citet{shenfeld2025rlsrazor} also highlights the benefit of on-policy data, in a manner complementary to ours, yet their hypothesis deviates slightly when instantiated in our setting as demonstrated in Appendix~\ref{app:base_policy_kl}.
% \nr{I wouldn't refer to Appendix A.5 from here. I think we should drop this sentence and refer to Appendix A.5 from the concurrent work paragraph in the related work section.}

%\nr{I refactored the preamble and added a note on the advantage estimate/specific form of policy gradient update. We should also briefly mention this in Subsection 4.1 probably. Also, should we add here somewhere that the advantage estimate was proposed by prior work to be a cause?}
\subsection{On-Policy Data is the Primary Contributor for Mitigating Forgetting}\label{sec:on_policy_explains}

There are three distinctions between RL, as implemented via GRPO, and SFT (\emph{cf.}~\S\ref{sec:prelim}): \emph{(i)} RL trains on on-policy data, generated by the current policy, while SFT uses off-policy data; \emph{(ii)} the RL objective typically includes KL regularization with respect to the initial policy  while SFT does not; and \emph{(iii)} RL multiplies gradients of responses by an advantage estimate while SFT does not.
Below, we identify on-policy data (\emph{i.e.}, \emph{(i)}) as the source of RL's robustness to forgetting by ruling out the necessity of \emph{(ii)} and \emph{(iii)}.
We note that our results stand in contrast to \citet{lai2025reinforcement}, which hypothesized that a particular form of advantage estimation mitigates forgetting.

% The lesser degree of forgetting with RL is commonly attributed to algorithmic choices such as KL regularization and the advantage estimation \citep{lai2025reinforcement}. We show in this section that these factors are not the primary contributor to forgetting mitigation.

%\paragraph{The advantage estimator in GRPO does not underlie its robustness to forgetting.}\label{sec:rl_var}
%Concurrent work \citep{lai2025reinforcement} attributed the robustness of GRPO to forgetting to an implicit regularization of the advantage estimator.
%The fact that SFT on approximately on-policy data does not suffer from forgetting (as shown above) stands in contrast to this hypothesis.
%In Appendix~\ref{app:reinforce}, we provide further support for on-policy data, and not any particular choice of advantage estimate, being the main factor mitigating forgetting by demonstrating that RL without an advantage estimator (\emph{i.e.,} REINFORCE \citep{williams1992simple}) is also robust to forgetting.

\input{tables/reinforce_grpo}
\input{figs/raft_gain_drop}

\textbf{KL regularization does not explain robustness to forgetting.}
KL regularization is commonly applied during RL to prevent the policy from drifting too far from its initialization \citep{ouyang2022training, shaoshuai2024deepseekmath}. We examine whether this regularization accounts for the lesser forgetting of RL.
As Figure~\ref{fig:kl_ablations} shows, non-regularized GRPO achieves a similar target task gain and non-target tasks drop tradeoff as KL-regularized GRPO across all considered models and datasets, except for Llama models trained on IFEval.
These results suggest that the use of KL regularization does not underlie the robustness of RL to forgetting.

\textbf{REINFORCE can be as robust as GRPO to forgetting.}\label{sec:reinforce}
We compare GRPO with REINFORCE \citep{williams1992simple}, a classical policy gradient RL algorithm that does not employ an advantage estimator.
Table~\ref{tab:reinforce_grpo_performance} shows that REINFORCE lags behind GRPO in optimizing the target task accuracy, yet maintains a similar low level of forgetting.
This suggests that algorithmic differences, such as the advantage estimator used in RL, primarily affect the magnitude of performance gains, whereas the mitigation of forgetting can be primarily attributed to the use of on-policy data.

\subsection{Approximately On-Policy Data Can Suffice for Mitigating Forgetting}\label{sec:approx_on_policy}
% \subsection{Approximately On-policy Data Already Benefits Forgetting Mitigation}\label{sec:approx_on_policy}

We identified the usage of on-policy data as the main contributor for the robustness of RL to forgetting.
However, generating on-policy data at each step entails a non-negligible compute overhead.
We therefore explore the degree of ``on-policyness'' required to enjoy the benefit of reduced forgetting.
Figure~\ref{fig:gain_drop} showed that Self-SFT, which generates data only from the initial policy, suffers from severe forgetting.
On the other hand, RL, which generates data at every step and thus represents the most on-policy end of the spectrum, is robust to forgetting.
We now test whether Iterative-SFT, an \emph{approximately} on-policy approach that iteratively trains on data generated at the start of each epoch \citep{zelikman2022star, dong2023raft, xiong2025minimalist}, can suffice for mitigating forgetting.

% \paragraph{SFT using approximately on-policy data.}\label{sec:iterative_sft}
%\nr{Need to rephrase this part since it already presumes on-policy data mitigates forgetting, while here the results are intended to both solidify this and show that approximate is enough.}
%How frequent should the on-policy data be generated for forgetting mitigation? 
%, whereas Expert-SFT at the opposite end uses fully off-policy data.
Figure~\ref{fig:raft_gain_drop} compares the target task accuracy and the drop in non-target tasks relative to GRPO of Iterative-SFT, Self-SFT, and SFT.
We find that Iterative-SFT is able to reach a target accuracy that is higher than or comparable to that of SFT, while only exhibiting mild to no forgetting.
% This demonstrates that, although training on data generated by the initial policy is not enough to reduce forgetting, approximately on-policy data sampled at each iteration enjoys similar benefits to fully on-policy data in terms of forgetting.
We also test an additional approximately on-policy approach that applies SFT on data generated during an RL run, and observe reduced forgetting as well (see Appendix~\ref{app:sft_on_rl_data}).
Overall, these results highlight that while RL remains most effective in forgetting mitigation, making SFT more on-policy or directly applying SFT on RL data can suffice for reducing forgetting.

%% file: tables/reinforce_grpo.tex
\begin{table*}[!t]
\centering
\resizebox{0.85\textwidth}{!}{%
\begin{tabular}{l l rr rr rr}
\toprule
 &  & \multicolumn{2}{c}{IFEval} & \multicolumn{2}{c}{MMLU} & \multicolumn{2}{c}{Countdown} \\
\cmidrule(lr){3-4} \cmidrule(lr){5-6} \cmidrule(lr){7-8}
      &        & Gain (\%) $\uparrow$ & Drop $\downarrow$ (\%) & Gain (\%) $\uparrow$ & Drop (\%) $\downarrow$ & Gain (\%) $\uparrow$ & Drop (\%) $\downarrow$ \\
\midrule
\multirow{3}{*}{Llama 3.1 8B Inst.} & SFT & 25.2 & 27.8 & 11.1  & 38.5 & 25.5 & 36.4 \\
& REINFORCE & 17.8 & 7.7  & 8.6  & -0.1 & 7.5  & -0.8 \\
                  & GRPO      & 18.4 & 3.4  & 14.6 & -0.2 & 60.4 & -0.5 \\
\midrule
\multirow{3}{*}{Qwen 2.5 7B Inst.} 
& SFT & 24.2  & 5.6 & 9.4 & 14.6 & 10.4 & 29.2 \\
& REINFORCE & 5.7  & 2.9  & 6.4  & -0.6 & 11.9 & -0.1 \\
                  & GRPO      & 17.0 & 0.2  & 8.4  & 0.2  & 29.2 & -0.3 \\
\bottomrule
\end{tabular}%
}
\caption{\textbf{The advantage estimate of GRPO is not responsible for its robustness to forgetting.} Comparison of SFT, REINFORCE, and GRPO. The SFT and GRPO results are taken from Figure~\ref{fig:gain_drop}.}
\label{tab:reinforce_grpo_performance}
\end{table*}

%% file: figs/raft_gain_drop.tex
\begin{figure*}[!t]
    \centering
    \includegraphics[width=0.92\linewidth]{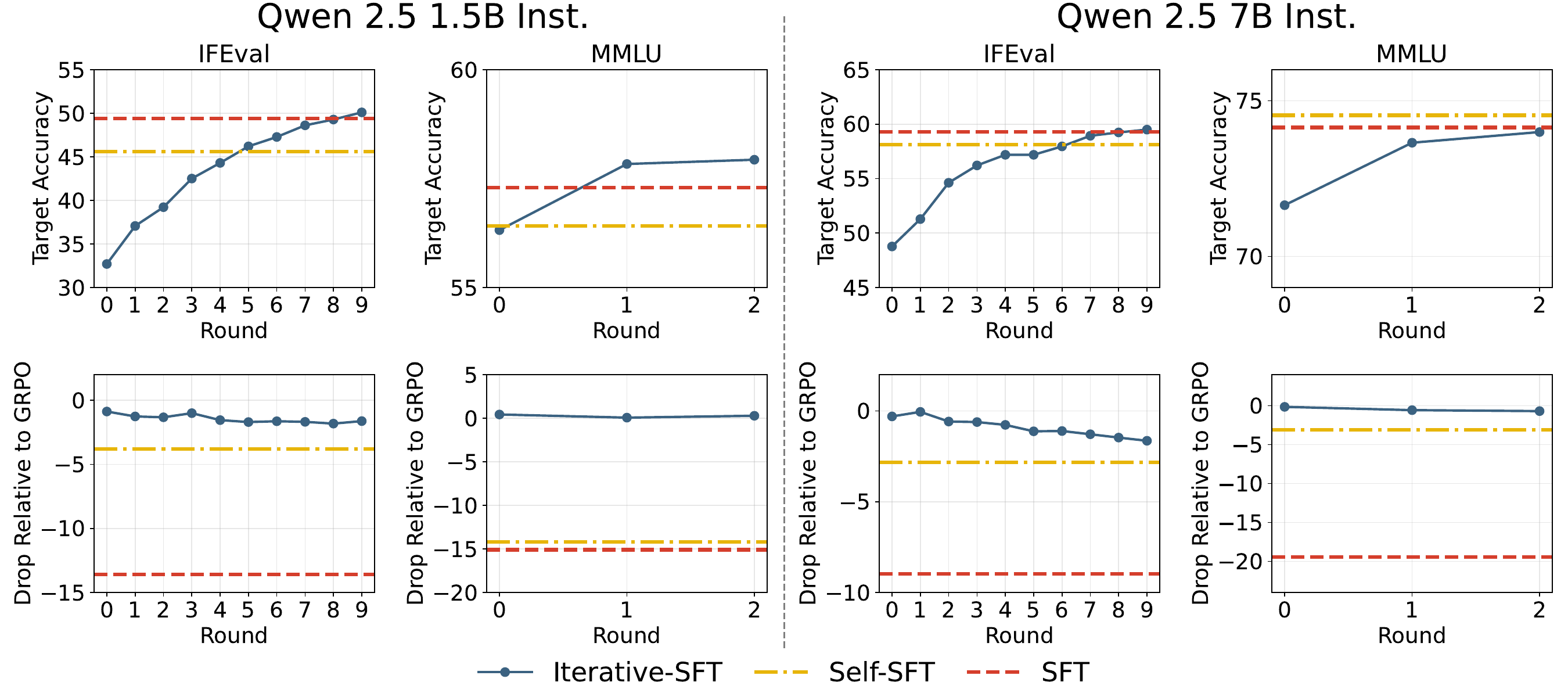}
    \caption{
      %\nr{I'd add a sentence here with the main takeaway from the plot, or rephrase the first sentence to also include the takeaway}
      %\nr{I'd refactor this caption by changing the takeaway sentence to just say what one should understand from the plot (Approximate on-policy data can suffice for mitigating forgetting in SFT), and only after add details in a couple of sentences or so on: (1) the figure comparing Iterative-SFT, Self-SFT, Expert-SFT and what each of these actually mean; and (2) how Self-SFT is not enough but Iterative-SFT is. Otherwise it is difficult to understand what this plot shows.}
      \textbf{Approximately on-policy data can suffice for mitigating forgetting in SFT.} 
      This figure compares three SFT variants using Qwen 2.5 1.5B Instruct and Qwen 2.5 7B Instruct on IFEval and MMLU. Top rows: target task accuracy. Bottom rows: average non-target accuracy drop relative to the drop of GRPO (i.e., $\Delta_d - \Delta_d^{\mathrm{GRPO}}$).
      The three SFT variants include: \emph{(1)} Iterative-SFT, which uses data generated at the start of each round (\emph{i.e.}, epoch); \emph{(2)} Self-SFT, which uses data generated from the initial policy; and \emph{(3)} SFT, which uses fully off-policy data generated by a separate expert model (Llama-3.3-70B-Instruct).
      % In all three variants, the generated data is filtered to only contain correct responses.
      Both Self-SFT and SFT appear as straight lines since they are trained for 2 epochs, and Iterative-SFT runs for multiple rounds until the target accuracy matches SFT.
      While SFT and Self-SFT suffer from severe forgetting, our results show that using approximately on-policy data, generated at the start of each epoch, can suffice for mitigating forgetting (Iterative-SFT).
      %, showing that per epoch on-policy data also mitigates forgetting.
      %Top: accuracy on the target dataset as a function of the number of SFT rounds (i.e., epochs). 
      %Bottom: average accuracy drop of the non-target datasets.
    }
    \label{fig:raft_gain_drop}
\end{figure*}

%% file: sections/related.tex
\section{Related Work}\label{sec:related}

\textbf{Catastrophic forgetting.}
Catastrophic forgetting has been studied since early research on connectionist models 
\citep{mccloskey1989catastrophic}. 
%Many early studies focus on simple neural networks with the focus of minimizing interference between tasks and prevent parameters to drastically change
Initial efforts to mitigate forgetting focused on preventing parameters from drastically changing
\citep{kirkpatrick2017overcoming, li2018learning, lopez-paz2017gradient}.
In the context of LM post-training, patterns of catastrophic forgetting differ due to the massive data used during pre-training \citep{luo2023empirical, shi2024continual, wu2024continual}. While LMs typically do not drastically forget all pre-trained knowledge, post-training LMs often leads to degradation in performance, which has been called ``alignment tax'' \citep{bai2022training, ouyang2022training}.
%In particular, \cite{chen2024continual} showed that training on knowledge with SFT leads to catastrophic forgetting, while much less forgetting observed when training on generic instruction following tasks.
Prior work found more severe forgetting when considering contrasting domains such as instruction following and safety \citep{qi2023finetuning, he2024what}.
Though, \citet{lee2024mechanistic, chen2024continual} suggest that forgotten behaviors or abilities can be revived with little re-training.
\cite{kotha2024understanding} posited that forgetting happens when the LM infers a wrong mode from the mixture of distributions to perform the task, and a carefully selected prompt can recover forgetting. Our work draws inspiration from the mixture-of-distributions perspective to establish intuition.

\textbf{LM post-training.}
Post-training methods such as SFT and RL are widely used for endowing pre-trained LMs with desired behaviors or enhancing performance on target tasks. 
SFT relies on the ground truth demonstrations.
By contrast, RL generates responses from the model and only provides a reward signal, be it from a parameterized reward model \citep{schulman2017proximal} or verifiable rewards \citep{shaoshuai2024deepseekmath,lambert2024tulu3}. 
Recent studies have shown that SFT and RL exhibit distinct characteristics.
\cite{razin2024vanishing,razin2025what} identified that the optimization speed of RL strongly depends on reward variance.
\cite{chu2025sft} showed that RL is able to generalize to unseen distributions while SFT can be prone to memorization. \cite{wang2025reinforcement} observed that RL can benefit from even training on a single example without severe overfitting.
Lastly, \cite{mukherjee2025reinforcement} reported that RL fine-tunes a smaller network compared to SFT.
A common thread connecting these results is that the parameter update during RL training is more local and targeted.
%RLVR \citep{lambert2024tulu3}.
%GRPO \citep{shaoshuai2024deepseekmath}.
%PPO \citep{schulman2017proximal}.
Other methods such as RAFT \citep{dong2023raft} and STaR \citep{zelikman2022star} perform SFT in several rounds and can be viewed as approximately on-policy RL.
This paper complements these studies and provides a forgetting-centric view on the difference between SFT and RL.

% \nr{Commented this part out since I do not believe it is super relevant}
% \textbf{Post-training induces pre-trained behavior.}
% Prior works have shown that SFT ``induces'' abilities that are inherent to the model as opposed to acquiring entirely new skill \citep{zhou2023lima}. Recently, similar phenomenon has been shown in RL where LMs can learn from answer format alone without correct ground truth label \citep{shao2025spurious} and lacking necessary pretrained knowledge hinders consequent RL training \citep{gandhi2025cognitive}. These observations indicates that it is necessary for pretrained data distribution to overlap with the post-training target ability.

%\hc{indicating that the old and new distributions are close}

%\textbf{Forward KL vs reverse KL.}
%Studies have connected SFT and RL through the lens of KL divergence.
%\cite{tajwar2024preference} showed that optimizing the policy gradient objective in RL can be viewed as minimizing the reverse-KL between the training policy and the optimal policy \citep{korbak2022rl}, which exhibits a mode-seeking behavior to fit the target distribution efficiently. In contrast, SFT typically uses the cross-entropy loss, which implements the forward KL objective
%This perspective has been used to preserve diversity in LM's generation \citep{li2025preserving}.

\textbf{Concurrent work.}
% Lai,shenfeld have also reported that RL is more robust than SFT to forgetting. Lai however suggest it is implicit regularization due to advantage estimate, whereas our experiments suggest otherwise and pinpoint to on-policy data. Shenfeld also highlight the benefits of on-policy data, but through a different, complemantary, perspective than the one we provide in Section 4. Furthermore, we explore the benefits of approximately on-policy data.
Similarly to our work, \citet{lai2025reinforcement,shenfeld2025rlsrazor} have concurrently found that RL exhibits less forgetting than SFT.
However, \citet{lai2025reinforcement} attribute RL's robustness to an implicit regularization of a particular advantage estimator.
We provide evidence against this claim in \S\ref{sec:analysis}, and instead identify the crucial role of on-policy data in mitigating forgetting.
\citet{shenfeld2025rlsrazor} also highlight the benefits of on-policy data through a perspective that is complementary to ours (\S\ref{sec:forgetting_dynamics}). 
%Despite the similar observations, their hypothesis deviates slightly when instantiated with more realistic datasets as opposed to the simplified setting, as we demonstrate in Appendix~\ref{app:base_policy_kl}.
Despite the similar observations, we find that their hypothesis on the connection between KL divergence from the initial policy and forgetting does not always hold in our setting (see Appendix~\ref{app:base_policy_kl}).
Moreover, our work goes beyond fully on-policy data and demonstrates the potential of approximately on-policy data in more efficiently mitigating forgetting.

%% file: sections/discussion.tex
\section{Conclusion}\label{sec:discussion}

%\nr{Requires some polishing and tweaking.}

We systematically compared catastrophic forgetting in SFT and RL for LM post-training. 
Across tasks, scales, and model families, we found that RL consistently achieves strong target performance with substantially less forgetting than SFT.
Our experiments in both simplified and real-world settings establish that the robustness of RL to forgetting primarily stems from its use of on-policy data, rather than other algorithmic choices such as the advantage estimate or KL regularization.
Furthermore, they highlight the potential of efficiently mitigating forgetting by incorporating approximately on-policy data, sampled asynchronously or at the start of each epoch.

\textbf{Limitations and future directions.}
Our work provides evidence that RL is more robust than SFT to forgetting across several tasks, model families, and scales.
However, investigating how forgetting patterns vary as the model and dataset sizes are further scaled, beyond our compute budget, remains a valuable direction for future work.
Moreover, while we provide intuition for why RL forgets less than SFT based on a simplified mixture-of-Gaussians setting (\S\ref{sec:forgetting_dynamics}) and empirically identify the use of on-policy data as a main cause for this difference in forgetting (\S\ref{sec:analysis}), additional research is necessary to theoretically establish the role of on-policy data in mitigating forgetting.
Going forward, the issue of forgetting becomes central as the community moves toward building agents that continually learn from experience \citep{silver2025era}.
Deciding what data to consume is consequential to the stability of the agent.
Our results indicate that data generated on-policy will better preserve existing capabilities, and is therefore safer to learn from, than off-policy data such as content on the internet or experience from other agents.
In a similar vein, our insights lays groundwork for understanding forgetting in the emerging paradigm of test-time training \citep{sun2020testtime, hardt2024tttnn}.

%% file: sections/ack.tex
\section*{Acknowledegment}\label{sec:ack}

We thank the members of Princeton Language Intelligence Group for providing comments on the manuscript. We also thank Kevin Lu from Thinking Machines Lab for the valuable feedback.
This research is supported by the National Science Foundation (IIS-2211779), Cisco Research, and is also supported in part by Schmidt Sciences.
NR is supported in part by the Zuckerman STEM Leadership Program.

%% file: sections/impact.tex
\section*{Impact Statement}
This paper investigates catastrophic forgetting in language model post-training. Our findings could help practitioners preserve safety behaviors and other desirable capabilities when fine-tuning models for new tasks, as we show that on-policy methods are more robust to forgetting than supervised fine-tuning. As language models are increasingly deployed as continually learning agents, understanding how training data choices affect capability retention has implications for building more stable and reliable AI systems. We do not foresee specific negative societal consequences beyond the general dual-use concerns common to research that improves language model training.

%% file: appendix.tex
\section{Additional Experiments and Implementation Details}\label{app:exp}

\subsection{Simulation Details}\label{app:simu_details}

\paragraph{Setup.} We use a univariate mixture-of-Gaussians synthetic task to compare forward KL (FKL; SFT analogue) update against reverse KL (RKL; RL analogue). 
We calculate gradient updates using $n=1000$ samples.
%At each training step we draw $n=1000$ samples, apply one FKL step using samples from the target new-mode. and one RKL step using REINFORCE. 
%To reflect a slow, pretrained mode, any update that affects the old-mode parameters is multiplied by a small factor (slow factor of $0.005$). 
For evaluation and plots, densities are computed on a uniform grid at every $100$ iterations. 
%For figures we cap the y-axis at $0.75$ and, when annotated, place labels near $(0.29,0.5)$ and $(0.69,0.5)$ in axes coordinates.

\paragraph{Single-mode setting.} 
We run the gradient step updates for $T=1000$ iterations or when the target task gain reaches $0.9$.
The training policy starts as a single-mode univariate Gaussian at the old mode, initialized as $\mathcal{N}(-3.2,1.0)$ ($75\%$ old mass), and is adapted toward the same target mixture used above: $0.75\cdot\mathcal{N}(-3.0,1.0)+0.25\cdot\mathcal{N}(3.5,0.7)$. We use an FKL learning rate $0.05$ and a RKL learning rate $0.05$.

\paragraph{Bi-modal setting.}
We run the gradient step updates for $T=1000$ iterations or when the target task gain reaches $0.9$. The initial policy $\pi_\theta(x)$ is a two-component mixture with weight $0.75$ on an ``old'' Gaussian $\mathcal{N}(-3.5,1.0)$ and $0.25$ on a ``new'' Gaussian $\mathcal{N}(0.5,0.7)$. The target $\pi^*(x)$ is a mixture with the same weights over $\mathcal{N}(-3.0,1.0)$ (old) and $\mathcal{N}(3.5,0.7)$ (new). We sweep two FKL learning rates $\{0.15,\,0.01\}$ and use a RKL learning rate $0.01$.

\input{figs/simulation_curves_3}

Figure~\ref{fig:simulation_curves_3} shows the bi-modal setting with forward KL at high and low learning rates and reverse KL. Forward KL with high learning rate results in catastrophic forgetting, yet forward KL with lower learning rate fails to learn to target mode. RL can cover the target mode without sacrificing the prior mode.

\paragraph{RL forgets when the target distribution is far.}

\input{figs/simulation_diff_dist}

We show in Figure~\ref{fig:simulation_diff_dist} the simulation results with varying distance ($4.0-6.0$) between $q_\mathrm{new}$ and $p_\mathrm{new}$. We observe that as the distance gets larger, RL begins to suffer from forgetting as well. suggesting that RL is not immune to forgetting when the target task is drastically far away from the starting modes.

%\hc{if the new mode is far away, even RL will forget $\rightarrow$ check public models}

\subsection{Connection Between Area Overlap and Total Variation Distance}\label{app:tv}

Let $f,g: \mathcal{D} \to \mathbb{R}_{\ge 0}$ be, possibly unnormalized, integrable density functions over a domain $\mathcal{D}$.
The total variation (TV) between $f$ and $g$ is defined by:
\[
\mathrm{TV}(f,g) \;:=\; \frac{1}{2} \int_{\mathcal{D}}  | f (y) - g (y) | dy 
\text{\,.}
\]
Notice that:
\begin{equation}
\begin{split}
\int_{\mathcal{D}} \min \left \{f (y) ,g(y) \right \} dy &= \int_{\mathcal{D}} \frac{1}{2} \big ( f(y) + g(y)-\lvert f(y)-g(y)\rvert \big ) dy \\
&= \frac{1}{2} \left(\int_\mathcal{D} f (y) dy + \int_{\mathcal{D}} g (y) dy\right) - \frac{1}{2}\int_\mathcal{D} \lvert f (y) -g (y) \rvert dy \\
&= \frac{1}{2} \left(\int_\mathcal{D} f(y) dy + \int_{\mathcal{D}} g(y) dy\right) - \mathrm{TV}(f,g)
\text{\,.}
\end{split}
\label{eq:min_int_tv}
\end{equation}

Now, in the context of \S\ref{sec:gaussian_uni}, recall that the \emph{area overlap} of the training policy $\pi_\theta$ with respect to the old mode of the optimal policy is defined by (Equation~(\ref{eq:uni_overlap_def})):
\[
S_\mathrm{old}(\theta) = \frac{\int_{-\infty}^\infty \min \left \{ \alpha^* p_\mathrm{old}(y), \pi_\theta(y) \right \} dy}{ \alpha^* } \text{\,.}
\]
Choosing $f = \alpha^* p_{\text{old}}$ and $g = \pi_\theta$, by Equation~(\ref{eq:min_int_tv}) we may write $S_\mathrm{old}(\theta)$ as follows:
\[
S_{\text{old}}(\theta)
=
\frac{\frac{1}{2}(\alpha^* +1)-\mathrm{TV} \big(\alpha^* p_{\text{old}},\pi_\theta \big)}{\alpha^*}
= 
\frac{1}{2}+\frac{1}{2 \alpha^*} - \frac{1}{\alpha^*} \mathrm{TV} \big(\alpha^* p_{\text{old}},\pi_\theta\big)
\text{\,.}
\]
Hence, the non-target tasks drop at training step $T$ is equal to the normalized increase in total variation distance between the training policy and the (scaled) old component of the optimal policy:
\[
\Delta_d  = S_\mathrm{old}(\theta_0) - S_\mathrm{old}(\theta_T) = \frac{ \mathrm{TV} \big(\alpha^* p_{\text{old}},\pi_{\theta_T} \big) - \mathrm{TV} \big(\alpha^* p_{\text{old}},\pi_{\theta_0} \big)}{ 
\alpha^*}
\text{\,.}
\]

Similarly, the area overlap of the training policy $\pi_\theta$ with respect to the new mode of the optimal policy is given by:
\[
S_{\text{new}}(\theta)
=
\frac{\frac{1}{2}(\alpha^* +1)-\mathrm{TV} \big(\alpha^* p_{\text{old}},\pi_\theta \big)}{\alpha^*}
= 
\frac{1}{2}+\frac{1}{2 (1 - \alpha^*)} - \frac{1}{1 - \alpha^*} \mathrm{TV} \big((1 - \alpha^*) p_{\text{new}},\pi_\theta \big)
\text{\,.}
\]
This implies that the target task gain at training step $T$ is equal to the normalized decrease in total variation distance between the training policy and the (scaled) new component of the optimal policy:
\[
\Delta_g  = S_\mathrm{new}(\theta_T) - S_\mathrm{new}(\theta_0) = \frac{ \mathrm{TV} \big((1 - \alpha^*) p_{\text{new}},\pi_{\theta_T} \big) - \mathrm{TV} \big((1 - \alpha^*) p_{\text{new}},\pi_{\theta_0} \big)}{ 
1 - \alpha^*}
\text{\,.}
\]

\subsection{Implementation Details}\label{app:training_details}

\paragraph{Training details.}   We used the AdamW optimizer. The learning rate was initialized to $1e{-4}$ for Llama-3.2-1B-Instruct and Qwen-2.5-1.5B-Instruct and  $5e{-6}$ for Llama-3.1-8B-Instruct and Qwen-2.5-7B-Instruct. We use cosine scheduler with warp-up step ratio $0.03$ over the course of training. 
Each model was trained with a batch size of $128$ for IFEval and MMLU and $64$ for Countdown.
Unless otherwise specified, training was run for $2$ epochs. 

For SFT, we minimized the cross-entropy loss with a maximum sequence length of $4096$. 
For Self-SFT, we generate $5$ responses from the initial model and filter out the incorrect responses based on the reward model. We keep all the correct responses of each prompt into the dataset.
For RL experiments, we used the GRPO algorithm with a KL-penalty coefficient of $0.05$ and apply updates right after the group samples are generated (hence no advantage clipping). We generate a group size of $5$ for each prompt.
All experiments were implemented in PyTorch and trained on maximally 8 H100 GPUs with mixed-precision (bfloat16) training. 

\paragraph{Data.}\label{app:data}
For IFEval \citep{zhou2023instruction}, we use the dataset provided in the Tulu3 for post-training \cite{lambert2024tulu3}. We split the dataset into the training set containing $13,000$ and the evaluation set of $1,972$ examples.
For MMLU, we split the dataset \cite{hendrycks2021measuring} into the training set containing $12,000$ examples and the evaluation set containing $2,042$ examples.
For Countdown, we generate data following the procedure in \cite{tinyzero}. We split the dataset into the training set containing $10,000$ examples and 
the evaluation set contains $1,000$ examples.

\paragraph{Prompts.}
Throughout our experiments, we use the following chat format:

\texttt{```\\
User:\\
\{prompt\} \\
\\
Assistant:\\
\{response\}\\
```}

For SFT, the target response is provided; for RL, the response is generated by the training policy at each optimization step.

\textbf{IFEval}: we use the prompt in the original dataset exactly.

\textbf{MMLU}: the prompt is:

\texttt{```\{question\}\\
Answer options:\\
A. \{option\_A\}\\
B. \{option\_B\}\\
C. \{option\_C\}\\
D. \{option\_D\}\\
\\
Reason about it and answer with "The answer is: <option>"```
}
, where the question and options are provided in the MMLU dataset.

\textbf{Countdown}: we use the following prompt
\texttt{"In this task, you need to use a list of numbers x = \{x\} to create an equation that leads up to the target number y = \{y\} using the basic arithmetic operations (+, -, *, /), and each number can only be used once. Think and return the final answer in \$\textbackslash\textbackslash boxed\{ \}\$."}
, where \texttt{\{x\}} is a list of integers and \texttt{\{y\}} is the target integer.

\subsection{Extra Experiments and Ablations}

%\subsubsection{REINFORCE is as Robust as GRPO to Forgetting}\label{app:reinforce}

%\input{tables/reinforce_grpo}

%In this appendix, we compare GRPO with REINFORCE \citep{williams1992simple}, a classical policy gradient RL algorithm that does not employ an advantage estimator.
%Table~\ref{tab:reinforce_grpo_performance} shows that REINFORCE lags behind GRPO in optimizing the target task accuracy, yet maintains a similar low level of forgetting.
%This suggests that algorithmic differences, such as the advantage estimator used in RL, primarily affect the magnitude of performance gains, whereas the mitigation of forgetting can be primarily attributed to the use of on-policy data.

%\subsection{Mixed Training and Replay}
%\hc{mixed training for SFT}
%\hc{revival of forgotten capability in SFT}

%\subsection{Dataset Examples}\label{app:dataset_examples}

%\paragraph{Is Online Update as Crucial as On-policy Data?}\label{sec:data_vs_update}
\subsubsection{SFT Using RL Traces}\label{app:sft_on_rl_data}
%\nr{I don't think this result is super important and looks like we are over 9 pages. Can briefly mention we also tested this and why this is interesting and refer to appendix for the full results.}
Data generated by RL throughout training is on-policy with regard to the model at each optimization step. 
When this RL data is later used for SFT, the process moves away from being fully on-policy, though it remains distinct from fully off-policy approaches such as SFT.
We investigate whether SFT on RL data can also mitigate forgetting.
%Utilizing RL's on-policy data for SFT removes it from the fully on-policy end of the spectrum, yet still distinct from the fully off-policy method such as Expert-SFT.
%We test if SFT on RL data also reduces forgetting.
In Figure~\ref{fig:sft_on_rl_data}, we observe that SFT trained on RL (GRPO) data trails full RL marginally in terms of gains but exhibits only slightly larger forgetting. 
%We have demonstrated in \S\ref{sec:iterative_sft} the crucial role of on-policy data for mitigating forgetting.
%However, RL algorithms not only leverage on-policy data but perform instant online update after each generation.
%To disentangle the importance of on-policy data and online update, we compare models trained using SFT on the GRPO generated data and GRPO.
This highlights a yet-to-be-identified benefit of using RL data for SFT \citep{deepseekr12025}.

\input{figs/sft_on_rl_data}

%\nr{The sentence below is a bit out of place, need to find a better place for it. Does it makes sense to have in Figure~\ref{fig:raft_gain_drop} also GRPO? Then this would be a good fit in the paragraph corresponding to that experiment.}
%This insight provides a practical guidance to balance across cost, performance, and forgetting.

%\paragraph{Bounds and special cases.}
%Since $\mathrm{TV}(f,g)\ge \frac{1}{2}\lvert \int f-\int g\rvert$, \eqref{eq:s-old-tv} and \eqref{eq:s-new-tv} imply $S_{\text{old}}(\theta),S_{\text{new}}(\theta)\in[0,1]$, with endpoints achieved in the obvious limiting cases. If $\lambda=1$ (single-mode ``old-only'' gold), \eqref{eq:s-old-tv} reduces to
%\[
%S_{\text{old}}(\theta)=1-\mathrm{TV}\!\big(p_{\text{old}},\pi_\theta\big),
%\]
%the standard overlap--TV complementarity for probability measures.

%\paragraph{Practical computation.}
%In our experiments all integrals are computed on a uniform grid via Riemann sums: the area overlap uses $\min\{f,g\}$ pointwise, and TV uses $\frac{1}{2}\lvert f-g\rvert$.

%\input{tables/ifeval_sft_on_rl}
%\input{tables/mmlu_sft_on_rl}
%\input{tables/ifeval_results}
%\input{tables/mmlu_results}
%\input{tables/countdown_results}

%\subsubsection{Off-Policy Data}
%\input{tables/swap}
%Table~\ref{tab:swap} further shows that the on-policy data only works for the model that generates it. Training on another model's on-policy data increases forgetting. 

% \subsection{$\mathrm{KL}[\pi_{\theta_0} \,||\,\pi_\theta]$ and Forgetting Correlates Moderately}\label{app:base_policy_kl}
\subsection{KL Divergence From Initial Policy and Forgetting Correlate Moderately}\label{app:base_policy_kl}

\begin{table}[ht]
\centering
\setlength{\tabcolsep}{6pt}
\begin{tabular}{ll*{6}{r}}
\toprule
\multirow{2}{*}{} & \multirow{2}{*}{} 
  & \multicolumn{2}{c}{\textbf{ IFEval }} 
  & \multicolumn{2}{c}{\textbf{MMLU}} 
  & \multicolumn{2}{c}{\textbf{Countdown}} \\
& & {Drop (\%)} & {KL} & {Drop (\%)} & {KL} & {Drop (\%)} & {KL} \\
\midrule
\multirow{3}{*}{Llama-3.2-1B-Instruct}
 & Self-SFT & 6.9 & 52.4 & 34.6 & 39.3 & 25.3 & 796.7 \\
 & SFT      & 26.2 & 61.1 & 28.9 & 48.4 & 24.2 & 1254.7 \\
 & GRPO     & 1.6 & 2.6  & 0.3 & 4.1  & -0.6 & 70.6 \\
\midrule
\multirow{3}{*}{Qwen-2.5-1.5B-Instruct}
 & Self-SFT & 3.0 & 25.0 & 14.0 & 3.0 & 19.5 & 896.1 \\
 & SFT      & 6.2 & 47.8 & 11.9 & 9.2 & 29.5 & 846.6 \\
 & GRPO     & 0.6 & 1.5  & 0.5 & 0.4 & 0.9 & 34.9 \\
\bottomrule
\end{tabular}
\caption{
Non-target tasks drop and $\mathrm{KL}[\,\pi_{\theta_0} \,||\,\pi_\theta\,]$ for the Llama-3.2-1B-Instruct and Qwen-2.5-1.5B-Instruct models from Figure~\ref{fig:gain_drop}.
}
\label{tab:drop_vs_kl}
\end{table}

\citet{shenfeld2025rlsrazor} have concurrently found that RL exhibits less forgetting than SFT, attributing RL's robustness to an implicit regularization of RL toward policies with low KL divergence from the initial policy $\pi_{\theta_0}$. In particular, \citet{shenfeld2025rlsrazor} empirically identify $\mathrm{KL}[\,\pi_{\theta_0} \,||\,\pi_\theta\,]$ as an indicator for the extent of forgetting, mostly through an extensive evaluation on synthetic tasks.
We explore the connection between $\mathrm{KL}[\,\pi_{\theta_0} \,||\,\pi_\theta\,]$ and forgetting in the setup of \S\ref{sec:sft_vs_rl}, where the KL divergence is estimated based on $100$ examples from the evaluation set.
Table~\ref{tab:drop_vs_kl} shows that, in accordance with the hypothesis of \citet{shenfeld2025rlsrazor}, GRPO exhibits both smaller KL divergences and smaller drops in non-target tasks performance compared to the SFT variants.
Furthermore, the Pearson correlation between KL divergence and non-target tasks drop across all models, methods, and datasets is $0.52$.
Yet, when comparing Self-SFT and SFT, the relation between KL divergence and forgetting is less monotonic---a larger KL does not necessarily imply a higher degree of forgetting.
This indicates that the relationship between KL divergence and forgetting is still not fully understood.

\subsection{Forgetting Evaluation in the Conversation Domain}\label{app:chat_results}

\begin{table}[!th]
\centering
\small
\begin{tabular}{lccc ccc ccc}
\toprule
& \multicolumn{3}{c}{IFEval} & \multicolumn{3}{c}{MMLU} & \multicolumn{3}{c}{Countdown} \\
\cmidrule(lr){2-4}\cmidrule(lr){5-7}\cmidrule(lr){8-10}
Model & Initial & SFT & RL & Initial & SFT & RL & Initial & SFT & RL \\
\midrule
Llama-3.1-8B-Instruct & 41.6 & 23.5 & 43.1 & 41.6 & 11.6 & 41.2 & 41.6 & 0.0 & 42.0 \\
Qwen2.5-7B-Instruct   & 36.5 & 22.5 & 35.2 & 36.5 & 19.9 & 36.0 & 36.5 & 0.4 & 34.6 \\
\bottomrule
\end{tabular}
\caption{Performance on AlpacaEval after SFT vs RL on each target benchmark.}
\label{tab:sft-vs-rl-transfer}
\end{table}

We include extra experiments on AlpacaEval (conversational task) and show that the benefit of RL in mitigating forgetting is consistent with the main experiments in our paper.
In Table~\ref{tab:sft-vs-rl-transfer}, we show the different runs evaluated on AlcapaEval, a benchmark for chat-based evaluation. We report WR (win rate in \%) against GPT-4 following standard practice.
We observe that for both Llama-3.1-8B-Instruct and Qwen-2.5-7B-Instruct, SFT suffers much more after training on the three different target tasks. On the other hand, we observe almost no drop or even improved performance for Llama-3.1-8B-Instruct after RL training compared to the base policy WR, and only very mild drop for Qwen-2.5-7B-Instruct after RL. The results are consistent with the other evaluations reported in the paper.

%\subsection{LLM Use In the Paper}

%We use LLM (ChatGPT) to aid writing. In particular, after drafting a sentence we use the LLM to ``Make it read better.'' or ``Make it read more fluently.''
%We also use LLM to help format the references and find the right source and venues that they are published in. We validate the links and details provided by the model to the best of our ability.

%% file: figs/simulation_curves_3.tex
\begin{figure}[h]
    \centering
    \includegraphics[width=0.99\columnwidth]{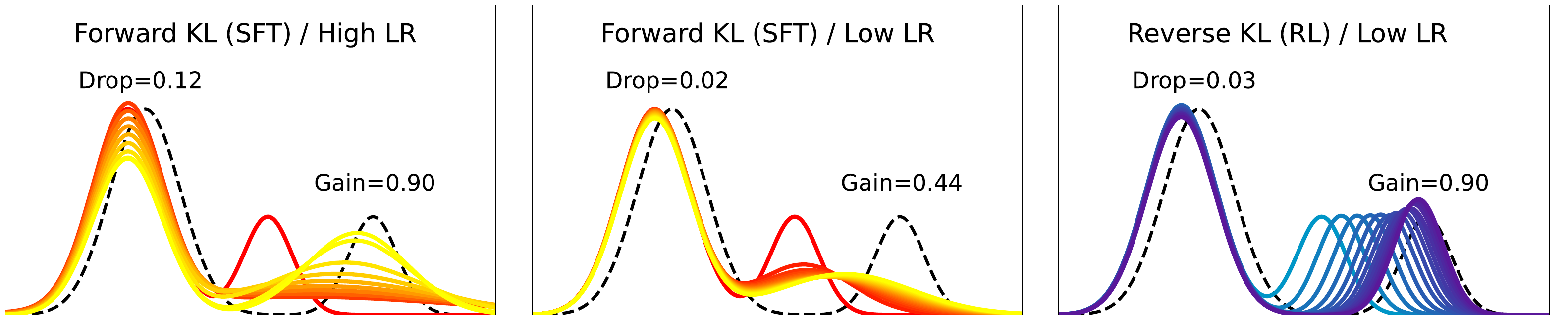}
    \caption{
        %\nr{I'd add a sentence here with the main takeaway from the plot, or rephrase the first sentence to also include the takeaway, and write in the titles Forward KL (SFT) and Reverse KL (RL), similarly to Figure~1}
        \textbf{Reverse KL (RL) with multi-modal training policy forgets less than forward KL (SFT).}
        Learning and forgetting patterns of forward KL with different high ($0.15$) and low ($0.01$) learning rates (left and middle) and reverse KL (right).
        Dashed lines represent the modes of the optimal policy: $p_\text{old}$ (left) and $p_\text{new}$ (right). 
        For forward KL, the data is sampled from the target mode $p_\text{new}$; the curve goes from red to yellow as training progresses. For reverse KL, the data is sampled from $\pi_\theta$; the curve goes from blue to purple.
        Forgetting corresponds to the decrease of overlap on the left mode and learning a new task is the increase in overlap on the right mode.
    }
    \label{fig:simulation_curves_3}
\end{figure}

%% file: figs/simulation_diff_dist.tex
\begin{figure}[h]
    \centering
    \includegraphics[width=0.98\columnwidth]{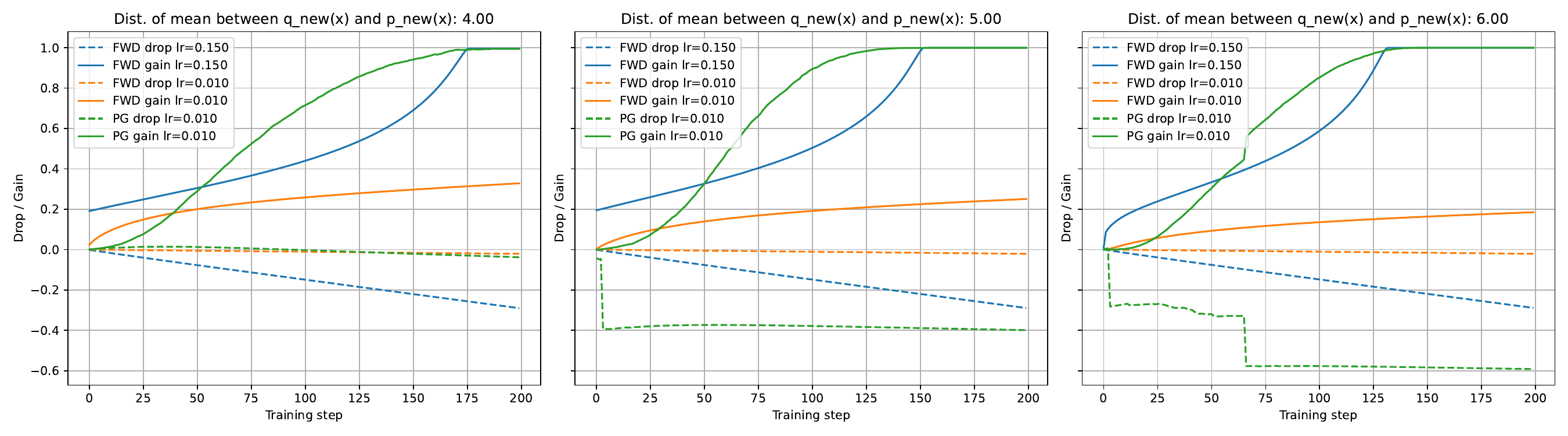}
    \caption{
        Simulation comparison with different distances ($[4.0, 5.0, 6.0]$) between $p_\mathrm{new}$ and $q_\mathrm{new}$.
    }
    \label{fig:simulation_diff_dist}
\end{figure}

%% file: figs/sft_on_rl_data.tex
\begin{figure}[!h]
    \centering
    \includegraphics[width=0.99\columnwidth]{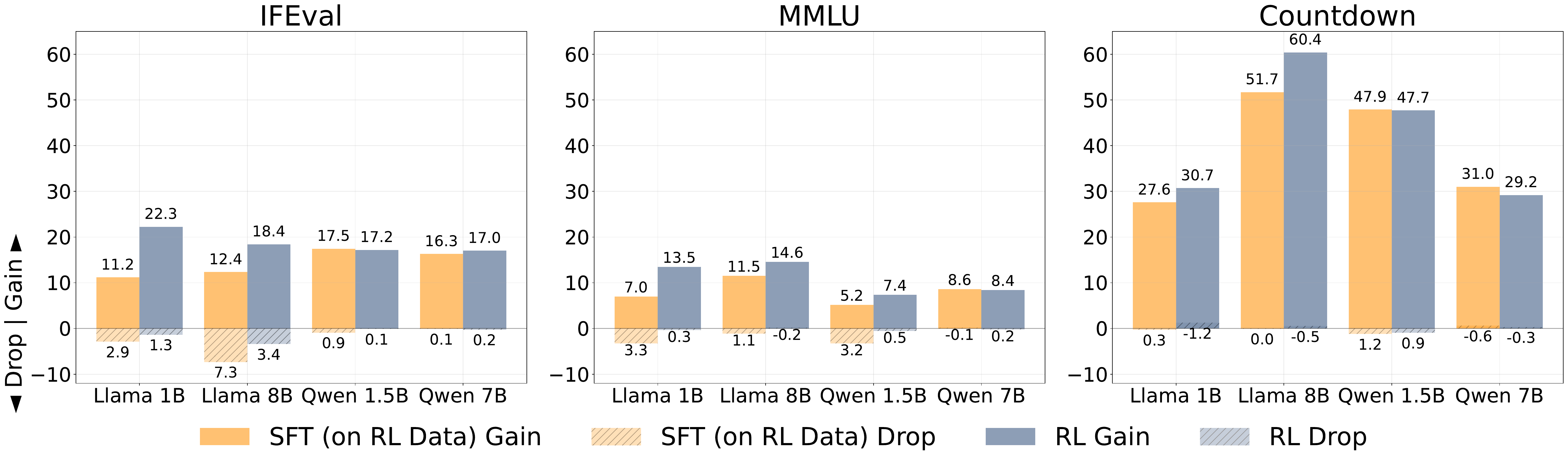}
    \caption{
        %\nr{I'd add a sentence here with the main takeaway from the plot, or rephrase the first sentence to also include the takeaway}
        %\nr{I'd change "SFT" in the legend to something else so that people do not confuse it with standard SFT. E.g., "SFT on GRPO data" or something of the sort.}
        \textbf{SFT over on-policy traces produced by RL exhibits reduced forgetting.}
        This plot shows the comparison between SFT trained on RL (GRPO) data and RL (GRPO).
        %\nr{\color{red}{TBH I'm not sure whether this result helps. I would recommend removing it or moving it to an appendix, since I don't think this contributes to our message or is an ablation that most people will wonder/care about. It can also be good to include here at least one other SFT variant so that it is easier to see SFT on GRPO data helped, despite still lagging behind GRPO.}}
        %\nr{\color{red}{Also, y-axis labeling is inconsistent with Figure~\ref{fig:gain_drop}}}
        }
    \label{fig:sft_on_rl_data}
\end{figure}